\documentclass[conference]{IEEEtran}
\usepackage{times}

\usepackage{graphics} 
\usepackage{epsfig} 
\usepackage{mathptmx} 
\usepackage{times} 
\usepackage{amsmath} 
\usepackage{amssymb}  
\usepackage{siunitx} 
\usepackage{textcomp}
\usepackage{gensymb} 
\usepackage[pagebackref=true,breaklinks=true,colorlinks]{hyperref}
\usepackage{booktabs}
\usepackage{multirow}
\usepackage{xcolor}

\let\labelindent\relax
\usepackage{enumitem}
\usepackage{xspace}

\usepackage{makecell}
\usepackage{multicol}
\usepackage{rotating}
\usepackage[percent]{overpic}

\usepackage{microtype}
\usepackage{contour}
\usepackage{courier}

\usepackage{subcaption} 
\usepackage[font=small]{caption}
\usepackage[percent]{overpic}

\usepackage[numbers]{natbib}
\usepackage{multicol}
\usepackage{enumitem}

\makeatletter
\DeclareRobustCommand\onedot{\futurelet\@let@token\@onedot}
\def\@onedot{\ifx\@let@token.\else.\null\fi\xspace}

\makeatother

\definecolor{MyDarkBlue}{rgb}{0,0.08,1}
\definecolor{airforceblue}{rgb}{0.36, 0.54, 0.66}
\definecolor{MyDarkGreen}{rgb}{0.02,0.6,0.02}
\definecolor{MyDarkRed}{rgb}{0.8,0.02,0.02}
\definecolor{MyDarkOrange}{rgb}{0.40,0.2,0.02}
\definecolor{MyPurple}{RGB}{111,0,255}
\definecolor{MyRed}{rgb}{1.0,0.0,0.0}
\definecolor{MyGold}{rgb}{0.75,0.6,0.12}
\definecolor{MyDarkgray}{rgb}{0.66, 0.66, 0.66}
\definecolor{MyPink}{rgb}{0.9, 0.33, 0.5}

\newcommand{\mypara}[1]{\par\vspace*{0mm} \textbf{{#1}}}

\def\PP{Pick\&Place}
\def\FLINGBOT{FlingBot}
\def\OURS{DextAIRity}
\def\OURSFLING{FlingBot+}
\def\OURSFIX{DextAIRity-fixed}

\def\SHAKE{Shake}

\begin{document}

\title{DextAIRity: \\Deformable Manipulation Can be a Breeze}
\author{Zhenjia Xu$^1$, Cheng Chi$^1$, Benjamin Burchfiel$^2$, Eric Cousineau$^2$, Siyuan Feng$^2$, Shuran Song$^1$ \\ $^1$ Columbia University \quad\quad\quad  $^2$ Toyota Research Institute\\ \href{https://dextairity.cs.columbia.edu/}{https://dextairity.cs.columbia.edu/}}



%

\maketitle

\begin{abstract}
This paper introduces DextAIRity, an approach to manipulate deformable objects using active airflow. In contrast to conventional contact-based quasi-static manipulations, DextAIRity allows the system to apply dense forces on out-of-contact surfaces, expands the system's reach range, and provides safe high-speed interactions. These properties are particularly advantageous when manipulating under-actuated deformable objects with large surface areas or volumes. We demonstrate the effectiveness of DextAIRity through two challenging deformable object manipulation tasks: cloth unfolding and bag opening. We present a self-supervised learning framework that learns to effectively perform a target task through a sequence of grasping or air-based blowing actions. By using a closed-loop formulation for blowing, the system continuously adjusts its blowing direction based on visual feedback in a way that is robust to the highly stochastic dynamics. We deploy our algorithm on a real-world three-arm system and present evidence suggesting that DextAIRity can improve system efficiency for challenging deformable manipulation tasks, such as cloth unfolding, and enable new applications that are impractical to solve with quasi-static contact-based manipulations (e.g., bag opening).
\end{abstract}

\IEEEpeerreviewmaketitle

\section{Introduction}
\label{sec:introduction}

\begin{figure}[t]
    \centering
    \vspace{2mm}
    \includegraphics[width=\linewidth]{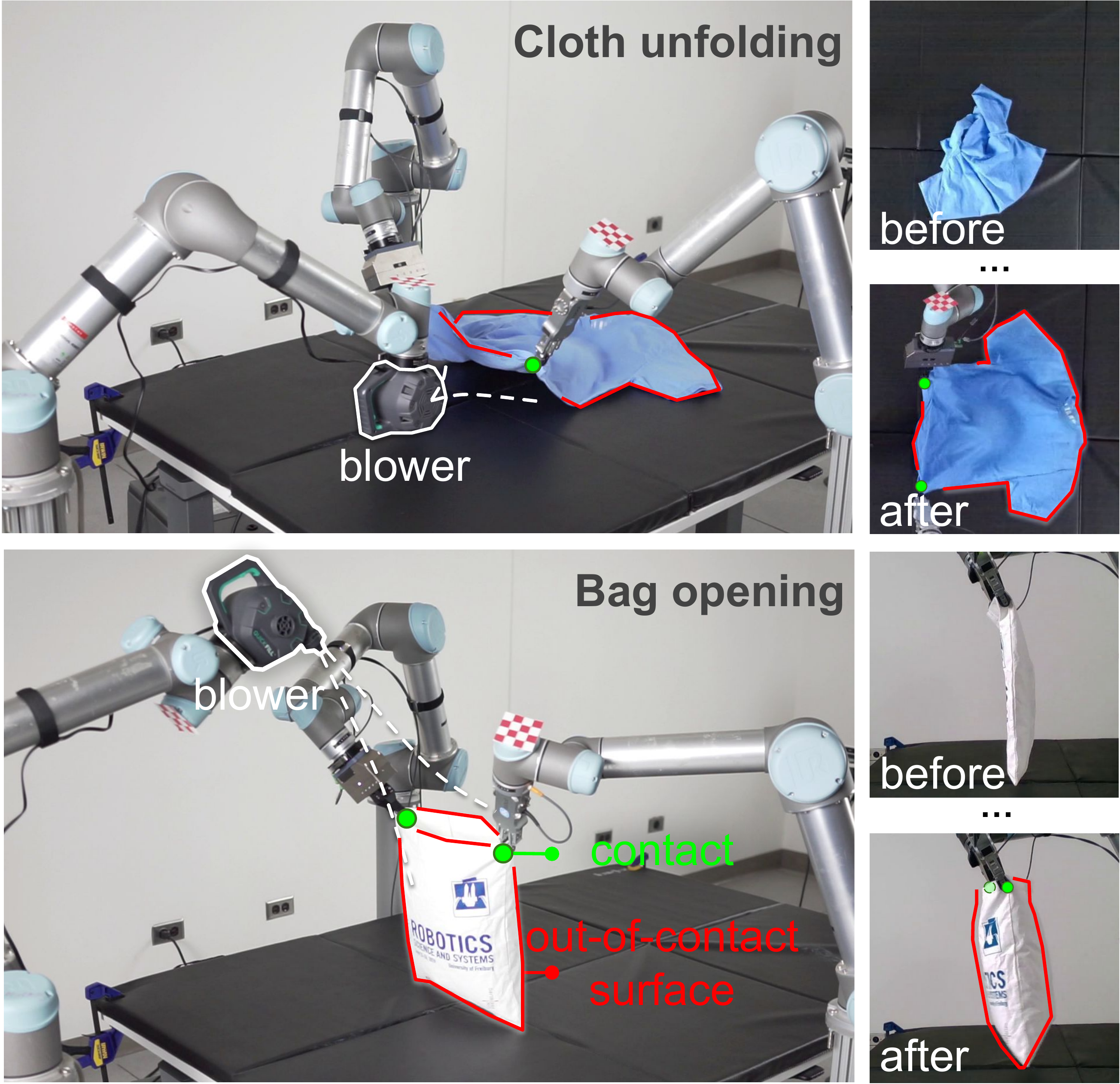}
    \caption{\textbf{\OURS} manipulates deformable objects by controlling an active airflow. We demonstrate \OURS~with two  tasks that are particularly challenging for traditional contact-based manipulation: unfolding a large piece of cloth (top) and opening a soft bag and maintaining its opened state (bottom). By controlling the blower's direction, the system can apply dense forces on out-of-contact surfaces (A and B) to efficiently achieve its goal. 
    }
    \label{fig:teaser}
    \vspace{-4mm}
\end{figure}

Many common everyday objects are impractical to manipulate via direct contact; however, they can be manipulated indirectly via air.  
From blowing leaves on the street to inflating molten glass, people purposefully control airflow to effectively change the state, such as pose or shape, of objects. 
In this paper, we seek to imbue robots with a similar capability, which we term \textbf{\OURS}. As a form of  contact-less dynamic manipulation, \OURS~provides a set of unique advantages over conventional contact-based quasi-static manipulation: 

\begin{itemize}[leftmargin=3.5mm]

\item \textbf{Dense force application.} Instead of applying force through \textit{sparse} contact positions, \OURS~allows the system to \textit{simultaneously} apply dense forces to a 3D space.
This property is particularly beneficial for under-actuated objects -- including deformables -- since it allows a robot to apply forces to those out-of-contact surfaces (e.g., Fig. \ref{fig:teaser} A,B). As a result, systems that are under-actuated with contact-based manipulation can become more controllable when manipulated via streams of air.

\item \textbf{Expanded workspace.} Since \OURS~does not require direct contact to manipulate objects, it can apply forces to objects that are distant from the robot and effectively expand its workspace. This property particularly is useful when the target object has a large volume or surface area -- spreading a large piece of cloth for instance -- and enables small robots to manipulate large items.

\item \textbf{High-speed interactions without high-speed robots.} Since the high-speed interactions are produced by the emitted airflow, these actions are much \textit{safer} than the actual robot movements at a similar velocity and easier to achieve without the need of high-end industrial hardware. 

\end{itemize}

\begin{figure*}[t]
    \centering
    \includegraphics[width=\linewidth]{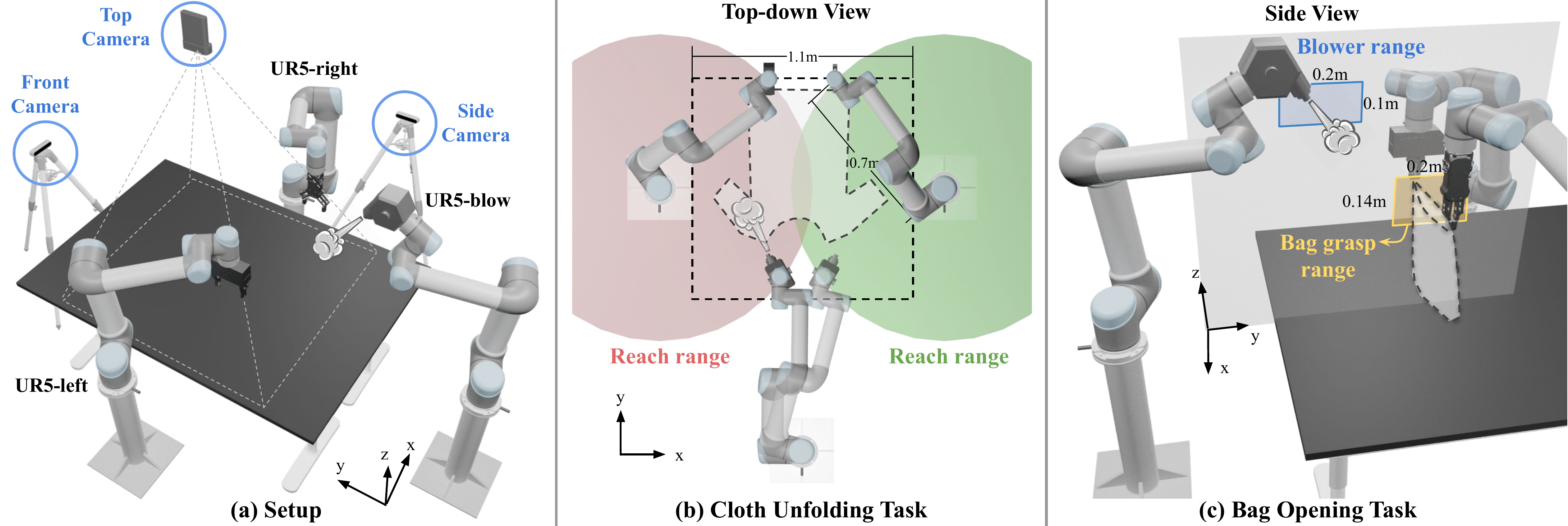}
    
    \caption{\textbf{System and Task Setup.} Our system setup consists of (a) three UR5 robot arms, two of which are equipped with parallel-jaw grippers and one with a commodity centrifugal air pump. (b) shows a top-down view of the workspace and robots' reach range for the cloth unfolding task. (c) shows a side view of the workspace and the robots' action space for the bag opening task. }
    \vspace{-2mm}
    \label{fig:setup}
\end{figure*}



However, despite the potential advantages of air-based manipulation, it is an open and challenging problem. First, accurately modeling the aerodynamics between the airflow and the environment is computationally costly, making model-based approaches impractical. 
Second, the precise airflow parameters, the shape and volume of an air-stream, are not easily observable or controlled, often resulting in highly stochastic dynamics and unexpected action effects. 

Both challenges motivate a self-supervised closed-loop solution for \OURS~that could learn and improve from data.  When airflow is applied to deformable cloth, the deformation on the cloth provides insight into the state information of the unobservable airflow. By combining this information with a closed-loop policy, the system can continually adjust its blowing action based on visual feedback.  Because the supervision signal for the learning algorithm can be directly computed from the visual observation, the system is fully trained with self-supervised trial-and-error --- without human demonstration or annotation.   
We demonstrate the effectiveness of \OURS~through two challenging tasks: (1) unfold a cloth to maximize its coverage and (2) open a bag to maximize its volume. 
For both tasks, we use a three-arm robotic system consisting of two arms with parallel grippers and one arm equipped with commodity centrifugal air pump (Fig. \ref{fig:teaser}).

The primary contribution of this work is to suggest a new approach for deformable object manipulation utilizing directed airstreams,  \OURS. 
Our simulation and real-world experiments suggest that \OURS~is able to improve efficiency for challenging manipulation tasks (e.g., unfolding an large pieces of unknown cloth) and enable new applications that were impractical with conventional contact-based manipulations (e.g., bag opening).   
We also discuss the potential limitations and necessary considerations of deploying \OURS~in real-world applications.  Code and robot videos: \href{https://dextairity.cs.columbia.edu/}{https://dextairity.cs.columbia.edu/}

\section{Related Work}
\label{sec:related_work}
 
\subsection{Quasi-static deformable object manipulation}

Manipulating deformable objects is a long-standing challenge in robotics. Methods have been developed for manipulating ropes \cite{sundaresan2020learning}, smoothing fabric \cite{seita2020deep,lin2021VCD}, folding cloth \cite{wu2019learning, lee2020learning, ganapathi2021learning, weng2021fabricflownet}, lifting bags  \cite{howard1996prototype, howard2000intelligent,seita_bags_iros_2021}, and inserting rigid object into deformable containers \cite{weng2021graph, seita2021learning}.
However, all the above methods are limited to using quasi-static pick-and-place actions. While this may be sufficient for rearranging rigid objects, this action space is generally inefficient when manipulating deformable objects. Since a deformable object has near-infinite degrees of freedom, and quasi-static actions can only be used to manipulate an object through contact (grasped area), these systems often require many, even hundreds of, interactions to achieve the goal. Tasks can sometimes be rendered impossible due to a robot's limited reach range and sparse contact positions (one contact point per gripper).

\subsection{Dynamic deformable object manipulation}
In contrast to quasi-static manipulation, dynamic manipulation \cite{mason1993dynamic} additionally leverages robot-produced acceleration forces to manipulate objects. This formulation allows the system to manipulate out-of-contact regions of the deformable object by building up an object's momentum with high-velocity actions \cite{wang2020swingbot,zhang2021robots,zeng2020tossingbot,casting,DensePhysNet}. 
For example, Ha and Song \cite{ha2021flingbot} propose a learning-based approach using high-speed flinging actions to unfold a severely crumpled cloth with greater efficiency than comparable quasi-static methods.
However, effectively using dynamic actions often come with restrictive system requirements, such as tracking the full (dense) state of cloth \cite{jangir2019dynamic} or using a high-speed camera and high-speed robots  \cite{balaguer2011combining, yamakawa2011dynamic, shibata2010robotic}. In many of these approaches, robots move with high acceleration using high controller gains, resulting in expensive hardware requirements and systems that pose a danger to things and people in their surroundings. Taken together, these attributes introduce substantial barriers to real-world deployment. In contrast, our approach is able to produce high-speed interactions using emitted air, which is inherently safer and inexpensive to implement. 

\subsection{Manipulation with air}
The idea of using airflow to generate directed force has been adopted in many aspects of robotic mechanical design. 
For example, airflow has been used in robot propulsion systems to produce thrust\cite{wasbari2017review}, as a pneumatic actuator to control a soft robot's geometry and articulation \cite{rus2015design, sanan2009robots}, or as a non-prehensile manipulator to move scattered objects \cite{wu2022learning}. 
Air manipulation is also used in many robot gripper designs, such as suction grippers \cite{zhakypov2018origami, yamaguchi2013development, bamotra2018fabrication} and non-contact grippers, using the Bernoulli principle \cite{erzincanli1998design, ozcelik2002non, ozcelik2005examination, davis2008end}.  
Early works also investigated systems that levitated rigid objects using controllable airflow \cite{nordine1982aerodynamic, tootchi2019modeling, escano2005position} including non-contact conveyance systems \cite{konishi1994conveyance, konishi1996two, konishi1999development, pister1990planar, biegelsen2000airjet}. 
In contrast to these prior approaches, which primarily focus on mechanical hardware design, our goal is to learn effective manipulation policies using off-the-shelf hardware (UR5 robots and a \$20 air pump). Moreover, instead of being limited to highly structured environments and objects with known physical properties, our algorithm is able to generalize and adapt to novel deformable cloths and bags using visual feedback.

\begin{figure}[t]
    \centering
    \includegraphics[width=\linewidth]{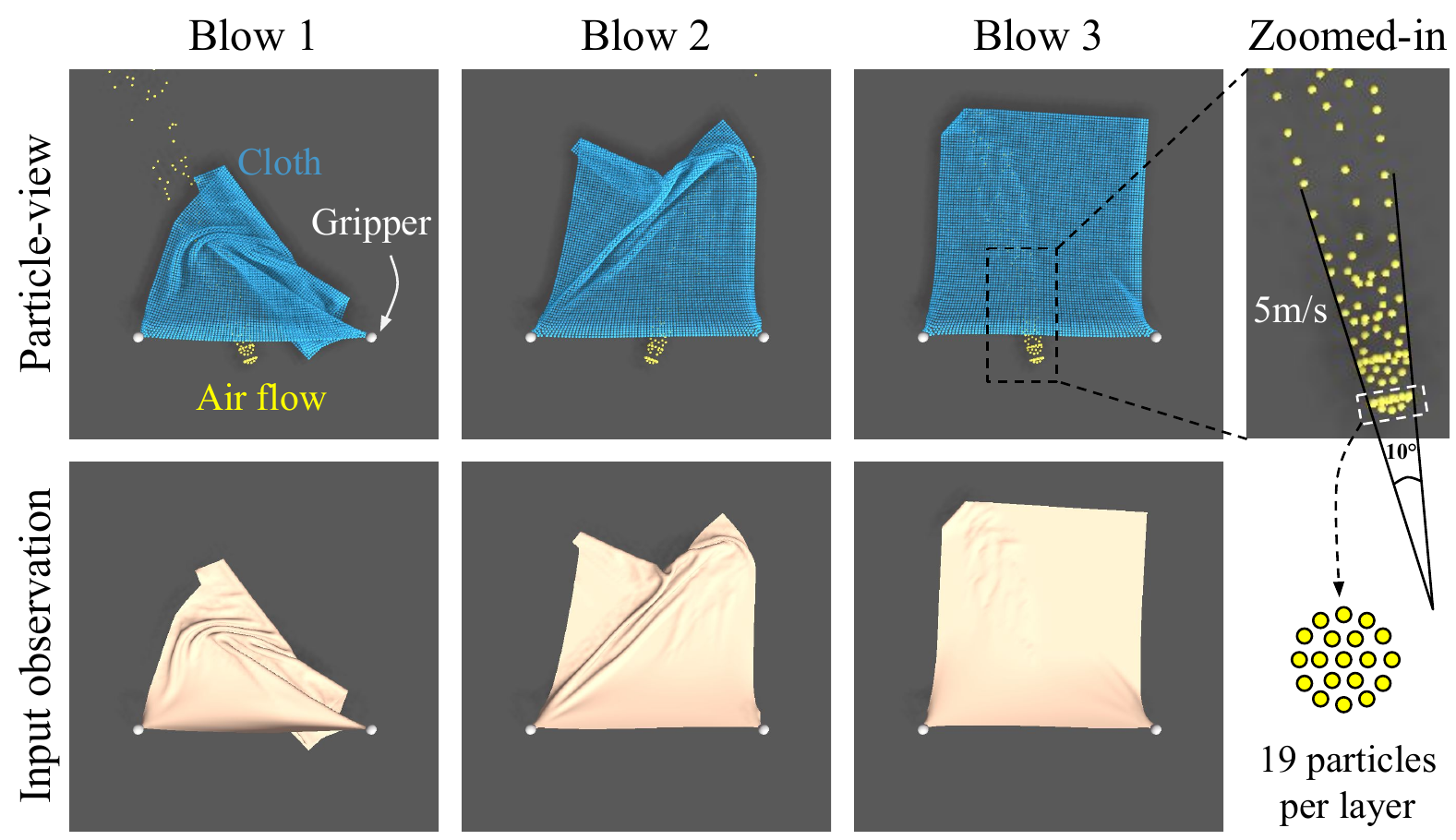}
    \caption{\textbf{Simulating Air-Cloth Interactions.} Cloth is simulated as a spring-mass system, and airflow is simulated as a stream of invisible particles. Our policy only takes the color image rendered by OpenGL as input. This simulation environment is only used for training cloth unfolding task but not bag opening task.} \vspace{-4mm}
    \label{fig:sim-env}
\end{figure}

\section{System and Task Setup}
\label{sec:method}
Our system setup consists of three 6-DoF UR5 robot arms, two of which are equipped with parallel-jaw grippers (a Schunk WSG50 and an OnRobot RG2), and one equipped with a commodity centrifugal air pump \footnote{Amazon link: \href{https://www.amazon.com/dp/B07F3S32TK}{https://www.amazon.com/dp/B07F3S32TK}.} The air pump is able to produce on-demand streams of focused air with a maximum flow rate of \SI{1100}{L/s}.
Our setup also includes three RGB-D cameras (highlighted in Fig. \ref{fig:setup}): a top-down Azure Kinect, a front-view Realsense D415, and a side-view RealSense D415. 

\subsection{Task Configurations} 

We use two tasks to examine the effectiveness of \OURS: one focuses on manipulating 2-D objects (unfolding cloth or garments), and the other extends to manipulating 3-D objects (opening a deformable bag). 

\textbf{Cloth unfolding:} The objective of this task is to increase the overall coverage of the cloth as measured by a top-down camera. This task is a typical first step for many cloth manipulation tasks such as cloth folding \cite{maitin2010cloth} or bed making \cite{seita2018bedmaking}. 
As shown in Fig. \ref{fig:setup}b, the workspace is defined as a $\SI{1.1}{m}\times\SI{1.1}{m}$ square in the x-y plane. A crumpled cloth is randomly dropped on the workspace and when fully unfolded can reach up to \SI{0.9}{m^2}. Because the UR5 robot has a \SI{0.7}{m} reach range, none of these robots can cover the entire workspace or the surface of a fully unfolded cloth. 
%

\textbf{Bag opening:} The objective of this task is to open a bag and maintain that opened state. It is a common first step in many downstream applications, such as filling a bag with objects \cite{seita_bags_iros_2021,seita2021learning}. The bag opening state is determined by thresholding the visible surface area (in pixels) of the bag observed by the side-view camera, where the threshold is selected for each bag to accommodate different bag sizes.
In our experiments, we assume that the closed bag is already grasped by two of the UR5 arms, and the grasping position in the y-z plane is uniformly randomly sampled in a 0.2m $\times$ 0.14m rectangle (the yellow region in Fig. \ref{fig:setup}c). To make the task more challenging, the distance between the two grasping points is randomly selected in $[l-\epsilon, l]$, where $l$ is the bag's maximum opening width and $\epsilon$ is \SI{0.05}{m}. The grippers can also tilt at a random angle, $\theta \in [\SI{0}{\degree}, \SI{30}{\degree}]$, along the x-axis to randomize the opening direction.
Similar to the cloth unfolding task, the policy infers blowing actions from top-down images in a closed-loop manner. We permit the execution of up to 4 blowing actions per episode before considering it a failure. 

\subsection{Simulation environment}
We use a simulated environment to train our cloth unfolding policy. 
The environment is built on top of the PyFleX bindings to Nvidia FleX \cite{ha2021flingbot,lin2020softgym,li2018learning}. In addition to simulating cloth and robot end-effectors, our simulation environment also provides a model of the blowing action. 
The blowing effect is simulated as a stream of invisible particles shot out from the blower. Air particles can be deflected from both cloth mesh and table. In each simulation step, 19 particles are uniformly shot out at \SI{5}{m/s} in a \SI{10}{\degree} cone shape (Fig. \ref{fig:sim-env}). The particles will then interact with the cloth using PyFleX's contact dynamics. Each blow lasts 150 simulation steps, where 150 is set empirically to make sure the cloth can reach to a relatively stable state (i.e., coverage fluctuates randomly within a small range).


Our particle-based wind effect has the key property of simulating non-uniform air effects on local regions of the cloth (for example, the targeted blown location). While PyFleX has a built-in ``wind'' option, it is only able to apply global force to objects and was not suitable for policy training.

It is important to note that our blowing simulation uses a heavily simplified physics model that does not reflect accurate aerodynamics, such as Bernoulli or Eddy effects, which are critical for applications like bag opening. Therefore, while we show the simulation is reasonable for training a cloth unfolding policy, it is not sufficient for bag opening tasks. As a result, we directly train the bag opening policy with real-world data.


\begin{figure*}[t]
    \centering \vspace{-1mm}
    \includegraphics[width=0.99\linewidth]{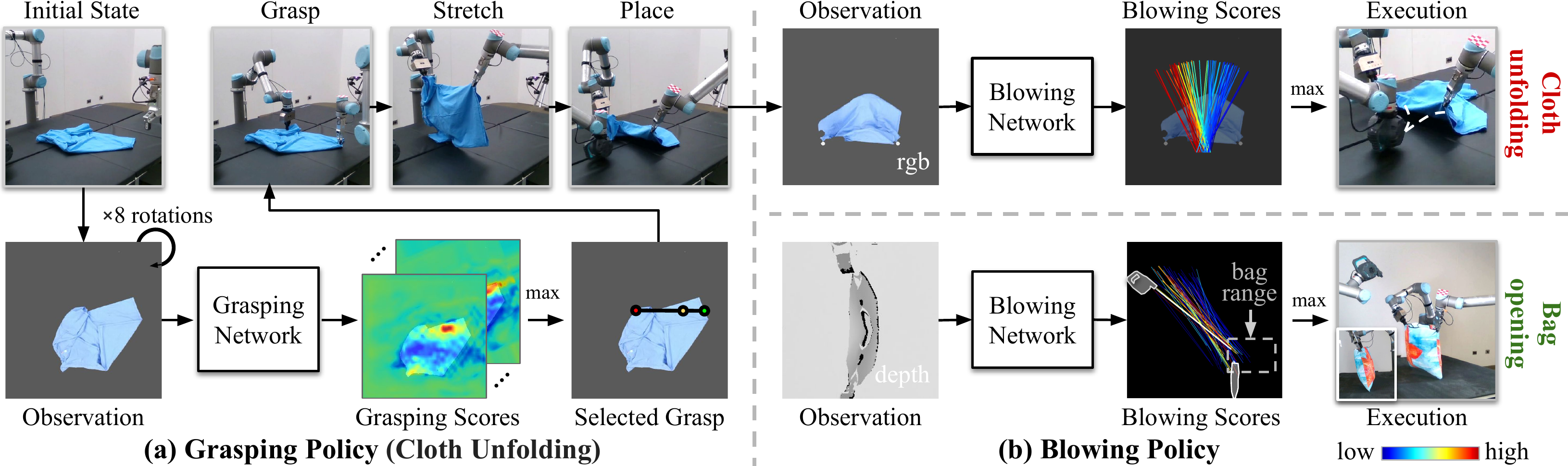}
    \caption{\textbf{Approach Overview.} (a) From a top-down observation, the Grasping Network predicts scores for each grasping action(i.e., center and rotation). The one with highest score will be selected for execution. Robots will grasp, stretch, and place the cloth down following blowing steps. (b) At each blowing step, the blowing network takes the top-down observation as input and infers blowing scores for each action candidate (blower position and rotation). The blowing action with the highest score will be executed.}
    \label{fig:approach} \vspace{-5mm}
\end{figure*}

\section{Method}

The key idea of \OURS~is to leverage the interactions between active airflow and deformable objects to achieve efficient manipulation. 
Crucially, while controlled airflow provides the system with additional control over out-of-contact object regions, the deformation on the object also provides visual feedback about the otherwise unobserved airflow, allowing the system to continually adjust its actions in a closed-loop manner. 
In the following section, we will first describe our method in the context of cloth unfolding. We will then present the required modifications for bag opening.

\subsection{Open-loop Grasping Policy}

To perform the cloth unfolding task, the system needs first to infer how to pick up cloth from the table. We extend the grasping framework from Flingbot \cite{ha2021flingbot} with a modified action parameterization to improve grasp quality. 

\textbf{Edge-coincident grasping action parameterization:} Flingbot used a dual-arm grasping action parameterized by a two end points on a line segment with center $C$, angle $\theta$, and width $w$. From these parameters, the two grasping positions $L,R$ are computed, allowing efficient computation grasping positions while satisfying collision-avoidance constraints.  
On top of this formulation, we further constrain the grasping positions $L, R$ to be on the edges of the cloth. We found this constraint significantly reduced the chance of grasping multiple layers of the fabric, which is a typical failure case for Flingbot. 
To implement this constraint, we directly extend the line segment (defined by $C$ and $\theta$) to intersect with the cloth mask, and two furthest intersection points are selected as the grasping positions ($L$ and $R$). The z-coordinate of these two grasping points are computed from depth. 
The object is segmented using background subtraction and if the selected center falls in the background, or the distance between two grasping positions is smaller than a minimum safety distance (\SI{0.1}{m}, defined empirically), no grasp will be executed, and the episode will terminate with a state reset.

\textbf{Grasping network:} To predict grasping parameters, we employ a spatial action map representation \cite{wu2020spatial, zeng2018learning, ha2021flingbot} to leverage the translational and rotational equivariance between the actions and the physical transformations of the cloth. The top-down color image is used to produce a batch of 8 candidates, differing by rotation about the z-axis, before being fed into the grasping network, which then outputs a grasping score for each pixel corresponding to predicted cloth coverage after execution. As a result, a pixel $C$ in the image rotated by $\theta$ directly maps to a corresponding grasping action. The grasping policy then selects the action associated with the highest predicted score for execution. We use DeepLabv3 \cite{chen2017rethinking} with random initialization as network architecture. We will describe training and supervision details in \S\ref{sec:method-training}.

\subsection{Closed-loop Blowing Policy}

After the the cloth is picked up, it is stretched taut using the front-view camera \cite{ha2021flingbot}. The robot will then move the cloth to \SI{10}{cm} above the table surface to prepare for execution of a blowing action. Initially the blowing is directed towards the center of the workspace and is then adjusted 4 times via a closed-loop blowing policy. Each blowing step lasts \SI{0.2}{s} after movement and the blower is kept on during all blowing steps.

\textbf{Blowing action parameterization:}
The blowing action is parameterized by position $\langle p_x, r_z \rangle$, where $p_x \in [\SI{-0.1}{m}, \SI{0.1}{m}]$ represents translation along the x-axis (left to right) and orientation $r_z \in [\SI{-30}{\degree}, \SI{30}{\degree}]$, where $r_z$ is a rotation angle around the z-axis (determining orientation in the x-y plane). Other blower parameters are fixed during the interaction -- the blower's nozzle is \SI{0.03}{m} above table and \SI{0.05}{m} away from the gripper holding position with a \SI{-10}{\degree} pitch angle.

\textbf{Blowing network:} To select effective blowing actions, we train a blowing network that takes the top-down color image and a blowing action as input, and predicts a score (i.e., the final coverage) for that action \cite{xu2022umpnet}. At each step, we uniformly sample $M=64$ blowing actions to evaluate and select the maximum one to execute (Fig. \ref{fig:approach}b, upper half).
The blowing network consists of an image encoder (7-layer convolution network) and an action encoder (3-layer MLP), followed by a 3-layer MLP to produce the final score. 

\subsection{Training procedure}
\label{sec:method-training}
Both the grasping and blowing networks are trained via self-supervised epsilon-greedy exploration. In each training episode, five grasping actions are executed, where each grasping action is followed by four blowing actions. Each blowing action is automatically labeled with the coverage of the cloth after execution, while each grasping step is supervised by observed coverage at the end of blowing execution. To compute coverage, the system obtains a cloth mask via background subtraction. Both networks are supervised via MSE Loss between predicted and real coverage.

Note that the performance of the two modules are highly coupled -- grasping score is dependent on the following blowing steps, and blowing performance is affected by how the cloth is grasped. This coupling can make training unstable. To solve this issue, we designed simple heuristic policies for grasping and blowing to allow independent pre-training for each module before combining them for further fine-tuning. The heuristic blowing policy is to place the blower in the middle of the workspace facing forward. Because this heuristic policy can unfold the cloth somewhat, it provides a reasonable starting place from which to bootstrap training. The heuristic grasping policy uniformly samples 100 grasping position pairs on the cloth and selects the pair with the largest distance. 
Both pre-training and fine-tuning are performed in simulation, which take 300 and 200 epochs respectively. Each epoch contains 32 episodes and 64 optimization steps with a batch size of 16 for the grasping network and 128 for the blowing network. Two FIFO replay buffers
(size=30000) are used to store training data. Both networks are implemented in PyTorch \cite{paszke2019pytorch} and trained using the Adam optimizer with a learning rate of 1e-4 and a weight decay of 1e-6.

\subsection{Modification for Bag Opening}
In the bag opening task, we assume the bag is already lifted up at a random position and the algorithm only need to infer the blowing actions. We use the same blowing network architecture as in the unfolding task, but with a few modifications in action parameterization, reward signal, and directly train this policy on real-world data. 

The blowing action is parameterized by $\langle p_y, p_z, r_x \rangle$, where ($p_y, p_z$) represents the blower's position in the y-z plane (the blue region in Fig. \ref{fig:setup}c), while $r_x$ corresponds to the pitch angle of the blower. At each blowing step, a top-down depth observation as input and the blowing action with the highest score will be executed. A binary reward, based on opening state, is computed by thresholding the surface area of the bag observed via the side-view camera. Note that the threshold is specific to each bag to accommodate different bag sizes. We choose to use binary reward for training due to our observation that, for this task using air, bags seldom exist in an intermediate state (half-open, for instance), and a binary reward is much more data-efficient for training. 
To reduce noise in reward computation, at the beginning of each blowing action the blower turns on for \SI{2}{s} after movement to ensure the bag is in a relatively stable state. Then, 400 images (20 fps) are captured by the side camera in the following \SI{2}{s} and the binary reward is determined by the averaged bag area over these images.
We directly train the blowing network with data collected by real-world robots via a random exploration. In total, we collected 4,400 (4,000 for training, 400 for validation) interactions over the course of 10 hours and trained the blowing network 50 epochs with a standard Cross Entropy loss.


\section{Evaluation} 
\label{sec:evaluation}
We evaluate \OURS~on cloth unfolding \S\ref{sec:eval-cloth} and bag opening tasks \S\ref{sec:eval-bag}. For both tasks, we evaluate task completion rate and ability to generalize to unseen cloths and bags on a real-world robot platform.  

\begin{figure}[t]
    \centering
    \vspace{2mm}
    \includegraphics[width=\linewidth]{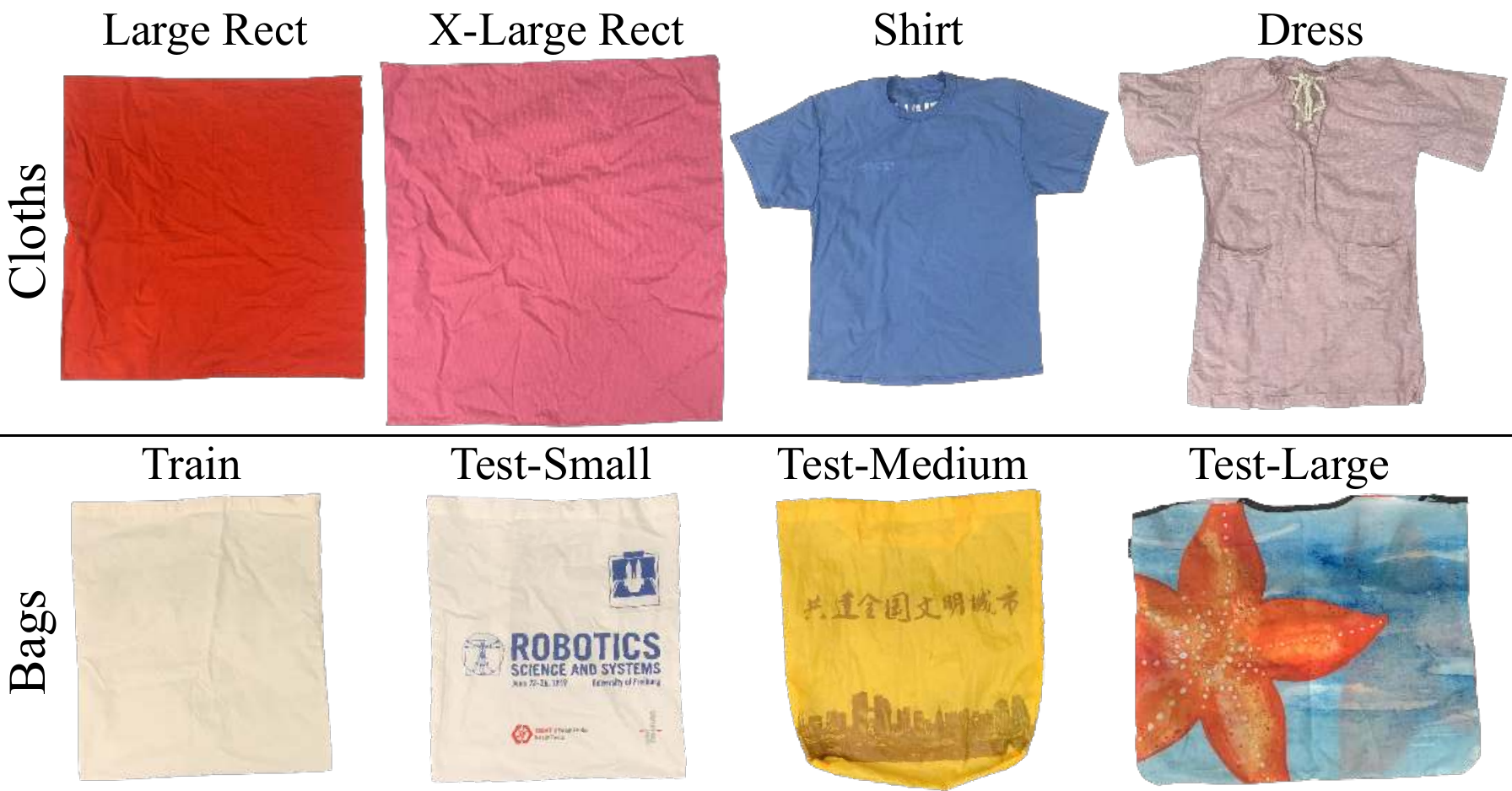}
    \caption{\textbf{Cloths and Bags used in Real-world Experiments.}}
    \label{fig:real-object}
\end{figure}

\begin{figure*}[t]
    \centering
    \includegraphics[width=\linewidth]{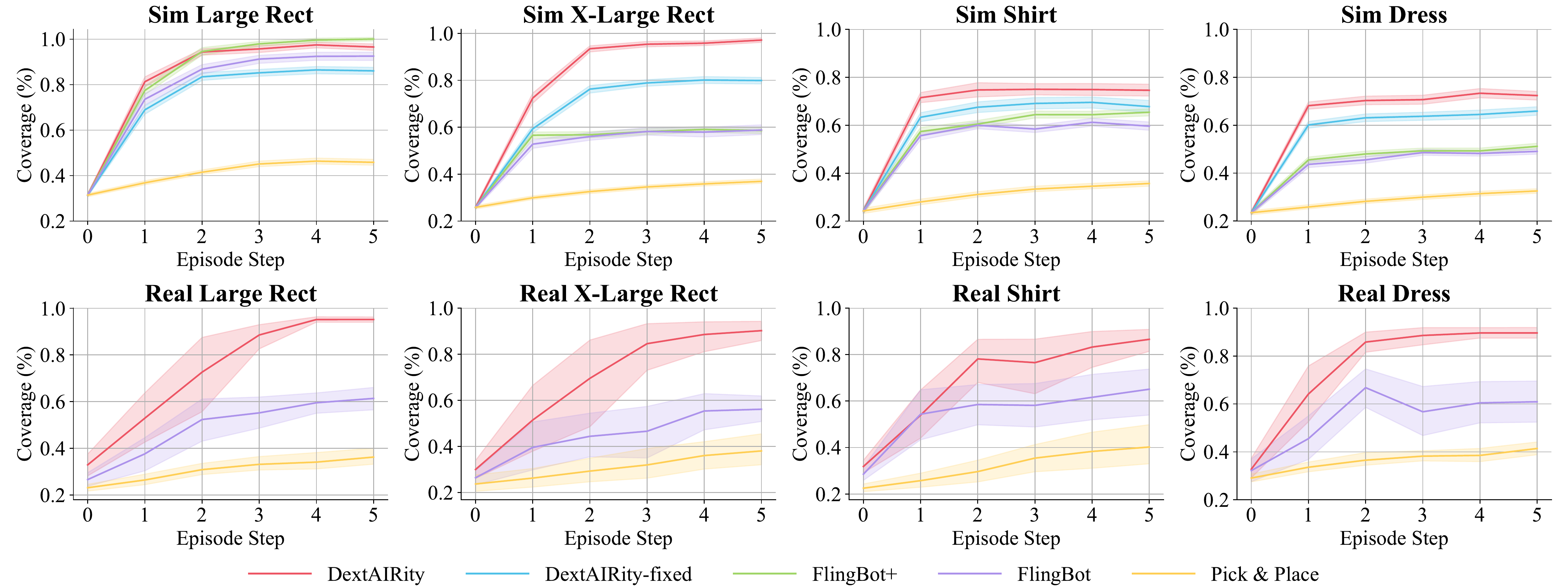}
    \caption{\textbf{Cloth unfolding coverage v.s. steps.}}
    \label{fig:cloth-curve}
    \vspace{-2mm}
\end{figure*}

\subsection{Cloth unfolding}
\label{sec:eval-cloth}
Performance is measured by cloth coverage at the end of each episode. The coverage statistic is normalized by the maximum possible coverage of the cloth in a manually flattened configuration. 
Each episode contains at most 5 interaction steps, where each step includes both grasping and blowing actions, and the policy terminates an episode when it predicts a pivot position outside the cloth mask. 

\mypara{Simulation Task Generation:} We generate five tasks for training and evaluation in simulation: 
\begin{itemize}
    \item {(Train) \textbf{Normal Rect}} contains rectangular cloths that are smaller in size than the robot's reach range. Edge lengths are uniformly sampled from \SIrange{0.4}{0.7}{m}. 
    \item {(Test) \textbf{Large Rect / X-Large Rect}} contain rectangular cloths with at least one side larger than the reach range. Edge lengths are uniformly sampled from \SIrange{0.4}{0.75}{m} for Large, \SIrange{0.8}{0.95}{m} for X-Large.
    \item {(Test) \textbf{Shirts / Dresses}} contain a subset of shirt and dress meshes from the CLOTH3D dataset \cite{bertiche2020cloth3d}. The average areas are \SI{0.41}{m^2} and \SI{0.55}{m^2} for shirts and dresses respectively, which are significantly larger than the shirt meshes used in FlingBot \cite{ha2021flingbot} whose average area is only \SI{0.14}{m^2}.
\end{itemize}
To reduce sim2real gaps, non-textured cloths are colored randomly. Cloth mass is sampled from \SIrange{0.2}{2}{kg} and the stiffness of stretching, bending, and shearing is fixed at \SI{0.8}{kg/s^2}, \SI{1}{kg/s^2}, and \SI{0.9}{kg/s^2} respectively. To generate a severely crumpled initial configuration, the cloth is grasped at a random position, held at a random height between [\SI{0.5}{m}, \SI{1.5}{m}], and dropped on the table to settle    .

\mypara{Real-World Tasks:}
Fig. \ref{fig:real-object} shows pictures of the four testing cloths used in our real-world experiments:
\begin{itemize}
\item {Large Rect}, which is $\SI{0.70}{m}\times\SI{0.70}{m}$ and \SI{0.081}{m}.
\item {X-Large Rect}, which is $\SI{0.76}{m}\times\SI{0.81}{m}$ and \SI{0.131}{kg}.
\item {Tshirt}, which is $\SI{0.71}{m}\times\SI{0.86}{m}$ and \SI{0.200}{kg}.
\item {Dress}, which is $\SI{0.80}{m}\times\SI{0.97}{m}$ and \SI{0.217}{kg}.
\end{itemize}
Note that the cloth items used in this experiment are all larger than those in FlingBot \cite{ha2021flingbot}, where the largest rectangle cloth is $\SI{0.7}{m}\times\SI{0.4}{m}$ and t-shirt is $\SI{0.53}{m}\times\SI{0.64}{m}$. 

\vspace{2mm}
\mypara{Ablations:} We compared with the following systems: 
\begin{itemize}
    \item \textbf{\PP} \cite{lee2020learning}: predicts a single-arm grasping position and movement direction for {quasi-static} pick-and-place.
    \item \textbf{\FLINGBOT} \cite{ha2021flingbot}: predicts a dual-arm grasping action for a {dynamic} flinging primitive.
    \item \textbf{\OURSFLING}: improved \FLINGBOT~uses our grasping policy to generate edge-coincident grasps.
    \item \textbf{\OURSFIX}: uses a fixed open-loop blowing action directed at the center of the workspace).
    \item \textbf{\OURS}: the system uses a learned blowing policy to produce closed-loop blowing actions from visual feedback. This is the full non-ablated method we propose.
\end{itemize}
\begin{table}[t]
    \centering
    \setlength\tabcolsep{4.6pt}
    \begin{tabular}{l|cc|cc}
    \toprule
                & \multicolumn{2}{c|}{Rectangle} & \multicolumn{2}{c}{CLOTH3D} \\
                &  Large                & X-Large               & Shirt                 & Dress  \\
    \midrule
    \PP         & 45.9 / 14.4           & 36.9 / 11.1           & 35.7 / 11.5           & 32.5 / 9.1 \\
    \FLINGBOT   & 92.6 / 61.1           & 58.8 / 32.9           & 59.6 / 35.4           & 49.1 / 25.6 \\
    \midrule
    \OURSFLING  & \textbf{100.0 / 68.6} & 58.6 / 32.7           & 65.4 / 41.2           & 51.2 / 27.7 \\
    \OURSFIX    & 86.1 / 54.6           & 79.9 / 54.0           & 67.8 / 43.5           & 65.9 / 42.4 \\
    \OURS       & 96.6 / 65.1           & \textbf{97.1 / 71.3}  & \textbf{74.6 / 50.3}  & \textbf{72.4 / 48.9} \\
    \bottomrule
  \end{tabular}
    \caption{\textbf{Simulation unfolding results} (final / delta coverage).}\vspace{-4mm}
    \label{tab:cloth-sim}
\end{table}
\mypara{Comparison with contact-based manipulation.} Here, we compare [\OURS] with state-of-the-art manipulation methods using quasi-static [\PP] and dynamic [\FLINGBOT] actions. 
In [Large Rect], cloth size is within or slightly larger than the robot arm's reach range. [\OURS] and [\FLINGBOT] achieve similar performance (over 90\% coverage after 3 steps), while [\PP] achieves is only 45.9\%. 

We also investigate these approaches’ performance on cloth that is significantly larger than the robot's reach range. In Tab. \ref{tab:cloth-sim}, final coverage of [\FLINGBOT] drops significantly when dealing with X-Large Rect (58.8 \%), while [\OURS] can still achieve a very large coverage (97.1\%). The failure of [\FLINGBOT] is due to its limited move speed, which needs to be prohibitively high to fully unfold a cloth much larger than robot arm's reach range. However, such high velocity is often dangerous and may not be feasible for many robots (e.g., the UR5 has a maximum speed of ~1m/s). 
In contrast, airflow can easily gain a high initial speed and apply forces to surfaces that are far away, resulting in an efficient system that is still safe to operate around humans.  

Overall, we find that quasi-static pick-and-place actions are generally inefficient for cloth unfolding and, while dynamic actions such as flinging can drastically improve efficiency, however, existing approaches still poses significant limitations when handling large or heavy cloth. Our experimental evaluation suggests that \OURS~is a promising approach for quickly and efficiently unfolding for large cloth items without the need of high-speed movements and large robots.

\begin{figure*}[!t]
    \centering
    \includegraphics[width=\linewidth]{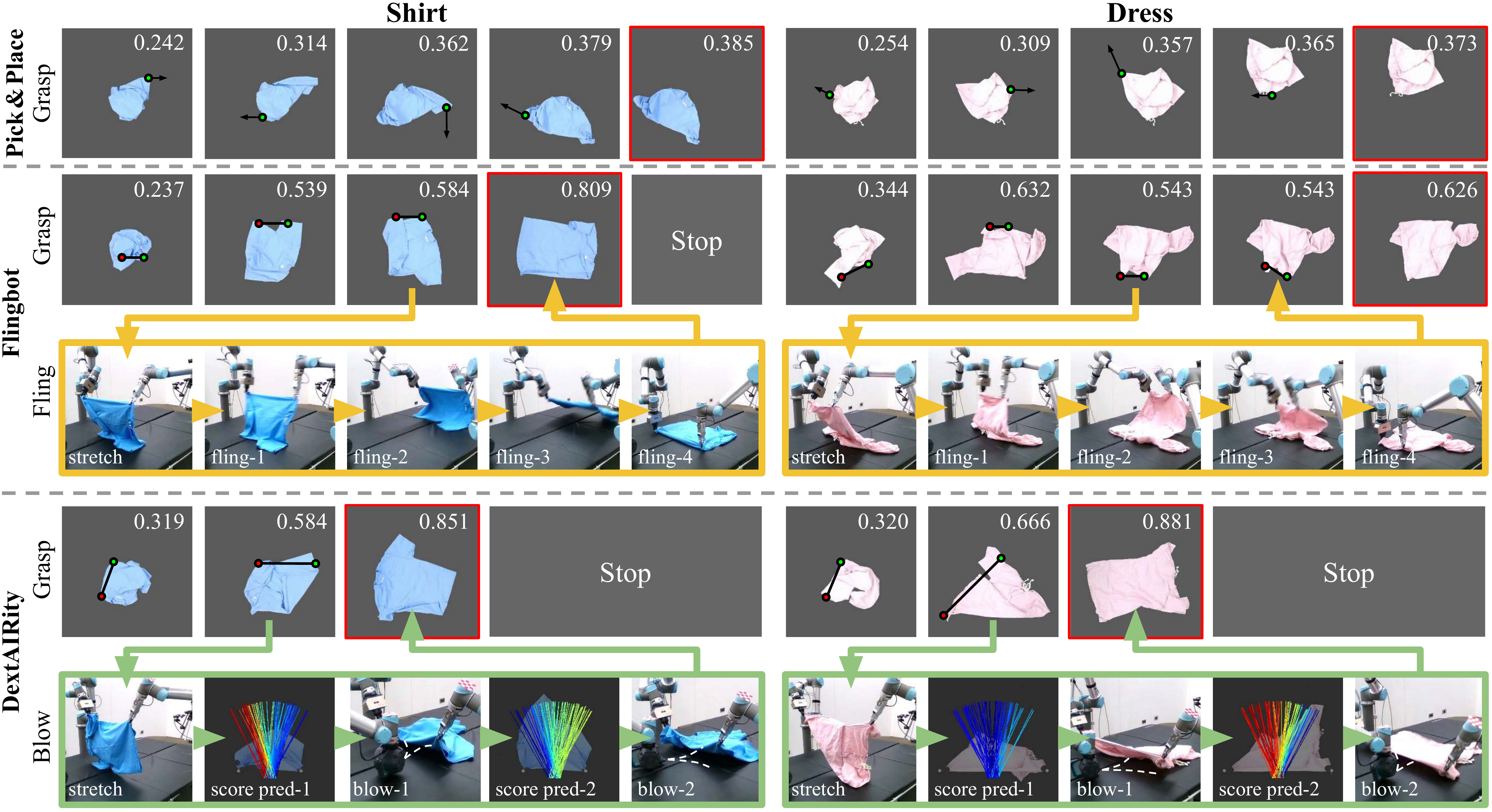}
    \caption{\textbf{Qualitative results of cloth unfolding.} Grasp predictions are visualized on the top-down image with cloth coverage labeled on the top right (row 1,2,4). The 3rd and 5th rows show the breakdown of fling and blowing actions respectively. \OURS~learns to grasp cloth corners and blow toward the unfolded part of the cloth to maximize coverage. While \FLINGBOT~discovered a similar grasping strategy, it can only half-unfold the long dress due to the speed constraints. Please see supplementary video for more results. } \vspace{-3mm}
    \label{fig:cloth-vis}
\end{figure*}

\textbf{Effectiveness of learned blowing policy.} Compared with [\OURSFIX], [\OURS] achieved higher final coverage(Tab. \ref{tab:cloth-sim}). Qualitative results in Fig. \ref{fig:cloth-vis} indicate that the blowing policy learns to detect and blow towards unfolded regions of the cloth, increasing effectiveness. As a result, coverage of X-Large Rect increases +23.0\%, +13.3\%, +1.6\%, and +0.3\% at each blow step compared to the fixed-policy ablation. In practice, while four blowing actions are executed after each grasping step, two were often sufficient.  

\begin{table}[t]
    \centering
    \begin{tabular}{l|cccc}
    \toprule
                  & Large Rect    & X-Large Rect  & Shirt         & Dress  \\
    \midrule
    \PP           & 36.2 / 13.1             & 38.0 / 13.3           & 40.2 / 14.7           & 41.4 / 12.3 \\
    \FLINGBOT     & 61.4 / 34.8             & 56.1 / 29.7           & 65.1 / 36.6           & 60.9 / 28.7 \\
    \midrule
    \OURS         & \textbf{95.2 / 62.3}    & \textbf{90.2 / 60.3}  & \textbf{86.6 / 54.7}  & \textbf{89.6 / 56.8} \\
    \bottomrule
  \end{tabular}
    \caption{\textbf{Real-world unfolding results} (final / delta coverage).}\vspace{-2mm}
    \label{tab:cloth-real}
\end{table}

\textbf{Benefit of edge-coincident grasp.} Compared with [\FLINGBOT], [\OURSFLING] achieves slightly better unfolding efficiency with the same fling primitive (Fig. \ref{fig:cloth-curve}a). Although the edge-coincident grasping policy provides a marginal improvement to final performance, it improves training efficiency significantly because the system no longer needs to learn the grasp width parameter. Training of [\OURSFLING] takes only 300 epochs, while [\FLINGBOT] requires over 2,000 epochs to converge. 

\textbf{Generalization to unseen cloth types} In this experiment, we investigate how well these approaches, trained on only rectangular cloths, generalize to novel cloth categories (shirts and dresses). Qualitative results in the real world (Fig. \ref{fig:cloth-vis}) suggest that even on out of distribution clothing, our learned grasping policy attempts to grasp cloth corners and the blowing policy preferentially directs air towards still crumpled regions of cloth. While this behavior is in-line with human intuition, we found the degree to which it generalized to novel cloth surprising. Similar to the in-distribution experiments, coverage plots in Fig. \ref{fig:cloth-curve} show the learned policy typically requires only 2 actions to reach the maximum coverage (around 70\% in simulation and 80\% in real).

\textbf{Evaluation on a physical robot.} We directly evaluate the trained model with our real-world setup. To promote policy transfer from simulation to reality,
we perform background removal and substitute a uniform-colored background consistent with the simulation environment. 
Tab. \ref{tab:cloth-real} shows performance averaged over 10 test episodes; our policy achieves over 80\% on all cloth types, outperforming [\FLINGBOT] and [\PP] by roughly 60\% and 40\% respectively. Our qualitative comparison, Fig. \ref{fig:cloth-vis}, suggests [\FLINGBOT] can successfully unfold shirts with width within the reach range but it fails (see the pink dress) when items become much longer. Even with the maximum fling speed, the dress can only be unfolded in half instead of fully unfolded.  In contrast, both cloths can be successfully unfolded via [\OURS] using a commodity air blower. The running time of these three primitives is 3.6s (blow$\times$4), 2.9s (fling), and 1.8s(place). With comparable execution time per step, [\OURS] achieves the best performance with fewer interactions. Please see our video for more results.

\begin{figure*}[!t]
    \centering
    \includegraphics[width=\linewidth]{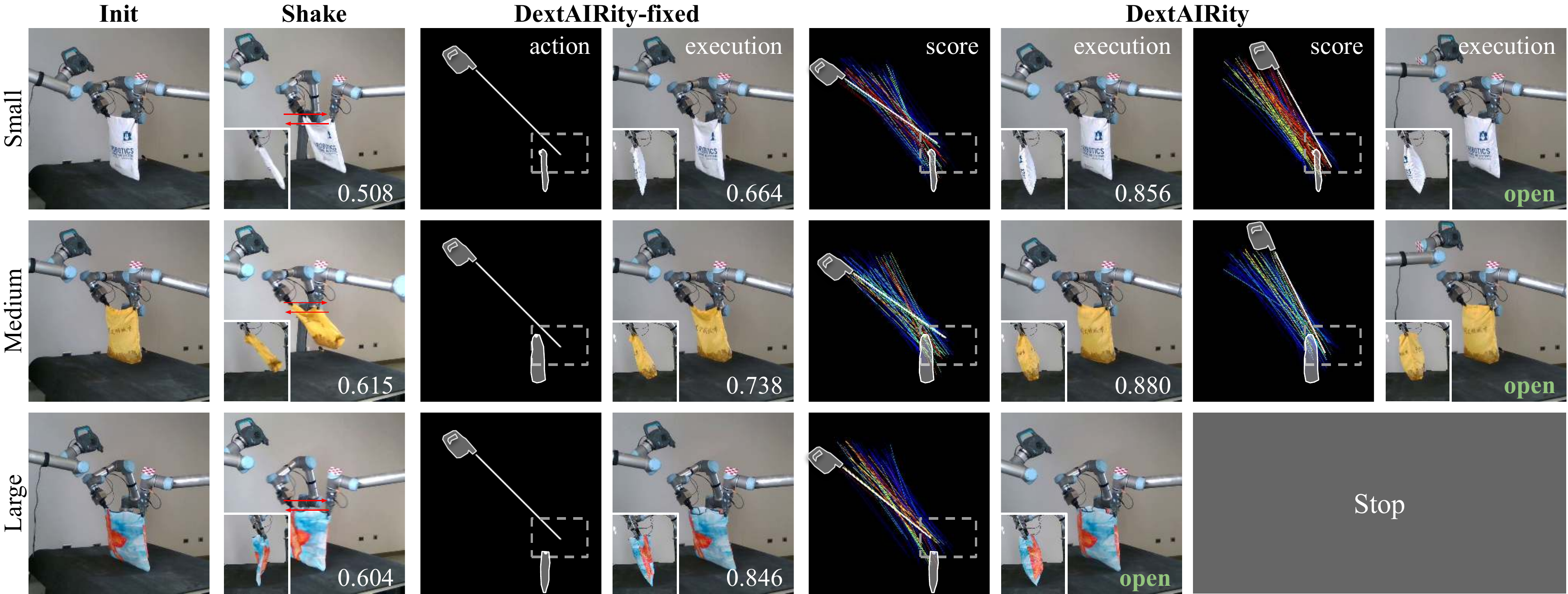}
    \caption{\textbf{Qualitative results of bag opening.} Bag state (normalized area if the bag is not opened) is labeled on the bottom right. Red arrows in Shake column indicate moving directions of end-effectors. Selected blowing actions are labeled as white lines. \OURS~generates blowing actions based on input observation and refines the action in a closed-loop manner. Please see supplementary video for more results.}
    \label{fig:bag-vis} \vspace{-3mm}
\end{figure*}

\begin{figure}[!t]
    \centering
    \centering
    \includegraphics[width=\linewidth]{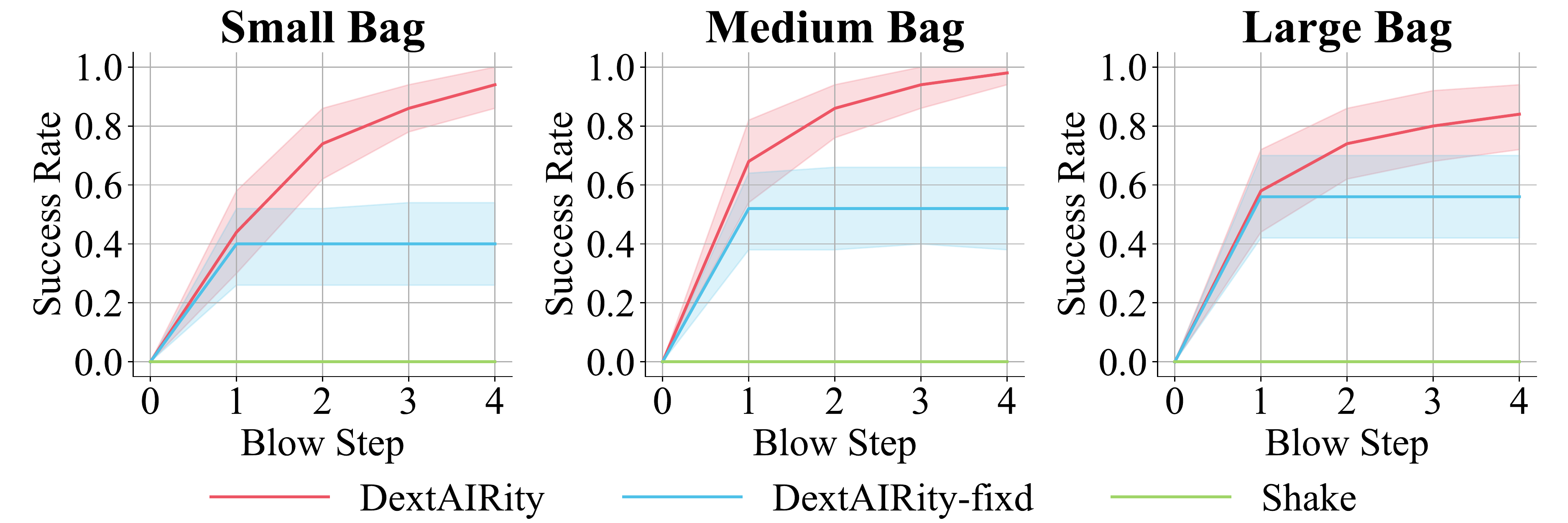}
    \caption{\textbf{Bag opening success rate v.s. steps.}}
    \label{fig:bag_result} \vspace{-1mm}
\end{figure}

\subsection{Bag opening}
\label{sec:eval-bag}
Task-performance for bag opening is measured by two metrics: 1) success rate: $\ p = \frac{1}{N} \sum_1^N sgn(A_i \geq \hat A)$, and 2) normalized bag area: $\bar A = \frac{1}{N} \sum_1^N min(A_i / \hat A, 1)$. $N$ is the number of testing cases, $A_i$ is the bag area of case $i$, $\hat A$ is the area threshold, and $sgn(\cdot)$ is the sign function. Note that the bag is considered open as long if its area is larger than the threshold, hence normalized area has a maximum value of 1.

\mypara{Ablations:} We compare our system with the following alternative approaches for bag opening:
\begin{itemize}[leftmargin=3mm]
    \item \SHAKE: moves the bag back-and-forth by rotating last joint and records the largest bag area during the process. 
    \item \OURSFIX: A fixed policy which blows toward the center of the workspace.
    \item \OURS: A closed-loop policy that predicts blowing actions based on  visual observations. In each episode, we run the policy 4 times or until the bag is opened.
\end{itemize}

\textbf{Training and testing bags:}
Fig. \ref{fig:real-object} shows our training and testing bags. We trained on a white cotton bag with size $\SI{0.37}{m}\times\SI{0.41}{m}$ and a \SI{0.050}{kg} mass; we evaluate the learned policy on three bags with very different size and materials: [Small] a cotton bag ($\SI{0.37}{m}\times\SI{0.41}{m}$, \SI{0.073}{kg}), [Medium] a plastic bag ($\SI{0.39}{m}\times\SI{0.43}{m}$, \SI{0.025}{kg}), and [Large] a plastic bag ($\SI{0.51}{m}\times\SI{0.42}{m}$, \SI{0.051}{kg}).

\textbf{Results}
Success rate and normalized area of both the training bag and novel testing bags are shown in Tab. \ref{tab:bag} and Fig. \ref{fig:bag_result}. We found that dynamic action [\SHAKE] generally fails to open the bag while [\OURSFIX] achieved a roughly 50\% success rate on the testing bags. In contrast, [\OURS] achieves 60\% success rate at the first interaction step and achieved a final success rate, after 4 blowing steps, of 88\%.

\begin{table}[t]
    \centering
    \begin{tabular}{l|cccc}
    \toprule
                   & Small Bag  & Yellow Bag          & Blue Bag  \\
    \midrule
    \SHAKE                        & 0.00 / 0.56                   & 0.00 / 0.68                   & 0.00 / 0.65 \\
    \OURSFIX                     & 0.40 / 0.86                   & 0.52 / 0.92                   & 0.56 / 0.94\\
    \OURS       & \textbf{0.98} / \textbf{0.99} & \textbf{0.94} / \textbf{0.99} & \textbf{0.84} / \textbf{0.99} \\
    \bottomrule
    \end{tabular}
    \caption{\textbf{Real-world bag opening} (succ rate / normalized area).}
    \vspace{-2mm}
    \label{tab:bag}
\end{table}


The improvement comes from two factors: 1) the information from visual observations (e.g. bag position and orientation) is utilized to infer a more effective blowing action (and explaining the performance gain during the first interaction step); 2) closed-loop manipulation allows the system to refine the blowing action adaptively  and compensate for errors in the learned model. 
Qualitative results in Fig. \ref{fig:bag-vis} show that [\OURS] tends to blow horizontally in the first step and adopts a more top-down blowing action in the subsequent steps. This strategy allows the system to first cover more space using a horizontal blow and collect observations on how the bag will react to the blowing action. Then, in the following steps, the policy will generate a more targeted, and thus more effective, blowing action based on these visual observations. 
In addition (similarly to our findings in the cloth-unfolding domain) our policy trained on the cotton bag was able to generalize to highly novel bags with different size and material. 

\begin{figure}[t]
    \centering
    \includegraphics[width=\linewidth]{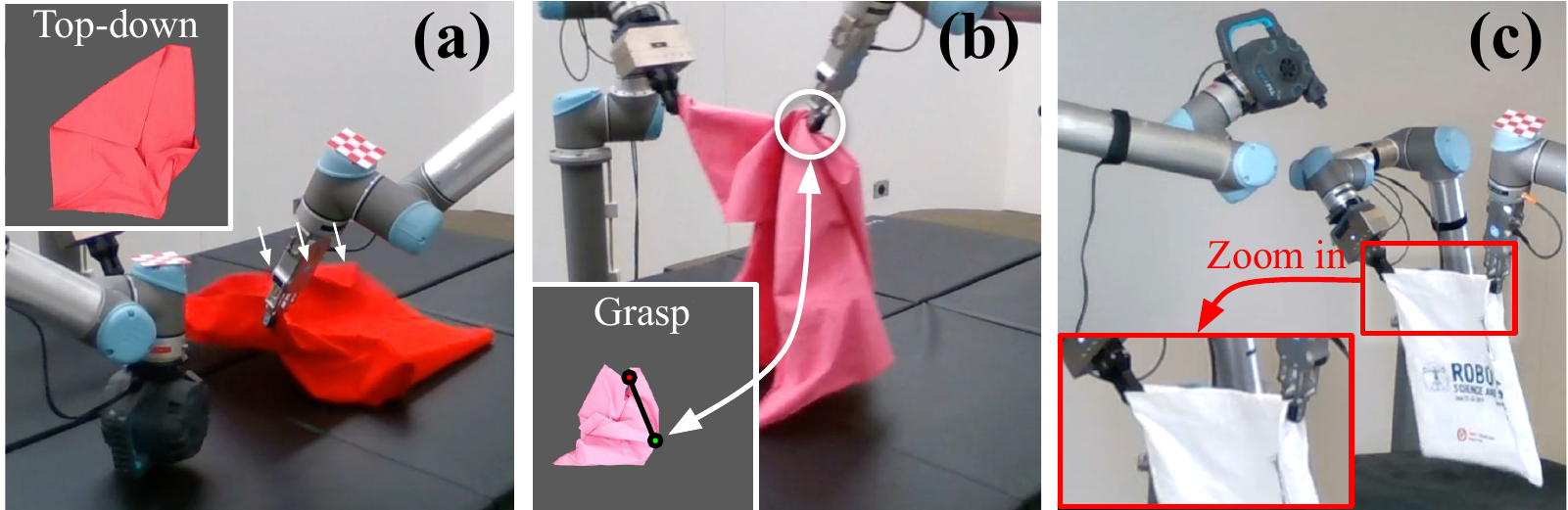}
    \caption{\textbf{Failure Cases.}  (a) A corner is inadvertently rolled up due to Eddy effects. (b) Multiple layers of the fabric are mistakenly grasped. (c) Two layers of the bag stick together and prevented air ingress.}
    \label{fig:failure} \vspace{-3mm}
\end{figure}

\section{Limitations and Practical Considerations}
While in this paper we demonstrate the effectiveness of directed air to manipulate deformable objects,  we discuss a few limitations and practical considerations of deploying \OURS~in real-world applications. 
First, during task execution, the air pump or blower is not particularly quiet. This issue could be mitigated with better sound isolation designs but could be problematic in some environments.
Second, commodity air pumps do not turn on and off instantaneously, which might be an issue for applications requiring fast impulse forces. 
Third, accurate aerodynamic simulation is non-trivial; depending on the target application, real-world data may be required to successfully train the system, which can be expensive and time-consuming to collect (though we do note that it would not generally require human annotation). 

In addition to these general constraints for air-based manipulation, Fig. \ref{fig:failure} shows a few typical failure cases of our specific system: 
a) In the unfolding task, the corner of a fabric can be rolled up by the air due to unmodeled aerodynamics effects, for instance eddy. This failure highlights the complexity of using aerodynamics as part of a manipulation strategy. 
b) While the edge-coincident grasp significantly reduced the frequency of grasping multiple layers of fabric, these poor grasps still occasionally occurred (Fig. \ref{fig:failure} b).
c) In the bag opening task, the bag can sometimes get in challenging states where two layers of fabric stuck together and prevented air ingress regardless of blow angle. To address this case, the robot would require a coordinated policy that simultaneously adjusts all three arms \cite{ha2020learning}, suggesting an exciting avenues for future explorations.

\section{Conclusion} 
\label{sec:conclusion}
We propose a new method for deformable object manipulation that utilizes active airflow, which we term \OURS. We demonstrate the effectiveness of \OURS~through two challenging deformable object manipulation tasks and we deployed the algorithm on a real-world three-arm system. Experiments suggest that \OURS~can improve system efficiency for challenging manipulation tasks like cloth unfolding and enable new applications that were not possible with quasi-static contact-based manipulations such as bag opening. We hope that the proposed algorithm, tasks, and results will help in spur the field to explore broader and more powerful forms of non-contact-based manipulation.


\section*{Acknowledgement} We would like to thank Huy Ha, Dale McConachie, Naveen Kuppuswamy for their helpful feedback and fruitful discussions. This work was supported by the Toyota Research Institute, NSF CMMI-2037101 and NSF IIS-2132519. We would like to thank Google for the UR5 robot hardware. The views and conclusions contained herein are those of the authors and should not be interpreted as necessarily representing the official policies, either expressed or implied, of the sponsors.

\bibliographystyle{plainnat}
\bibliography{references}

\begin{thebibliography}{53}
\providecommand{\natexlab}[1]{#1}
\providecommand{\url}[1]{\texttt{#1}}
\expandafter\ifx\csname urlstyle\endcsname\relax
  \providecommand{\doi}[1]{doi: #1}\else
  \providecommand{\doi}{doi: \begingroup \urlstyle{rm}\Url}\fi

\bibitem[Balaguer and Carpin(2011)]{balaguer2011combining}
Benjamin Balaguer and Stefano Carpin.
\newblock Combining imitation and reinforcement learning to fold deformable
  planar objects.
\newblock In \emph{2011 IEEE/RSJ International Conference on Intelligent Robots
  and Systems}, pages 1405--1412. IEEE, 2011.

\bibitem[Bamotra et~al.(2018)Bamotra, Walia, Prituja, and
  Ren]{bamotra2018fabrication}
Abhishek Bamotra, Pushpinder Walia, Avataram~Venkatavaradan Prituja, and
  Hongliang Ren.
\newblock Fabrication and characterization of novel soft compliant robotic
  end-effectors with negative pressure and mechanical advantages.
\newblock In \emph{2018 3rd International Conference on Advanced Robotics and
  Mechatronics (ICARM)}, pages 369--374. IEEE, 2018.

\bibitem[Bertiche et~al.(2020)Bertiche, Madadi, and
  Escalera]{bertiche2020cloth3d}
Hugo Bertiche, Meysam Madadi, and Sergio Escalera.
\newblock Cloth3d: Clothed 3d humans.
\newblock In \emph{European Conference on Computer Vision}, pages 344--359.
  Springer, 2020.

\bibitem[Biegelsen et~al.(2000)Biegelsen, Berlin, Cheung, Fromherz, Goldberg,
  Jackson, Preas, Reich, and Swartz]{biegelsen2000airjet}
David~K Biegelsen, Andrew~A Berlin, Patrick Cheung, Markus~PJ Fromherz, David
  Goldberg, Warren~B Jackson, Bryan Preas, James Reich, and Lars~E Swartz.
\newblock Airjet paper mover: An example of mesoscale mems.
\newblock In \emph{Micromachined Devices and Components VI}, volume 4176, pages
  122--129. International Society for Optics and Photonics, 2000.
\newblock URL \url{https://doi.org/10.1117/12.395620}.

\bibitem[Chen et~al.(2017)Chen, Papandreou, Schroff, and
  Adam]{chen2017rethinking}
Liang-Chieh Chen, George Papandreou, Florian Schroff, and Hartwig Adam.
\newblock Rethinking atrous convolution for semantic image segmentation.
\newblock \emph{arXiv preprint arXiv:1706.05587}, 2017.
\newblock URL \url{https://arxiv.org/pdf/1706.05587.pdf}.

\bibitem[Davis et~al.(2008)Davis, Gray, and Caldwell]{davis2008end}
S~Davis, JO~Gray, and Darwin~G Caldwell.
\newblock An end effector based on the bernoulli principle for handling sliced
  fruit and vegetables.
\newblock \emph{Robotics and Computer-Integrated Manufacturing}, 24\penalty0
  (2):\penalty0 249--257, 2008.
\newblock URL
  \url{https://www.sciencedirect.com/science/article/pii/S0736584506001347}.

\bibitem[Erzincanli et~al.(1998)Erzincanli, Sharp, and
  Erhal]{erzincanli1998design}
F~Erzincanli, JM~Sharp, and S~Erhal.
\newblock Design and operational considerations of a non-contact robotic
  handling system for non-rigid materials.
\newblock \emph{International Journal of Machine Tools and Manufacture},
  38\penalty0 (4):\penalty0 353--361, 1998.
\newblock URL \url{https://doi.org/10.1016/S0890-6955(97)00037-0}.

\bibitem[Escano et~al.(2005)Escano, Ortega, and Rubio]{escano2005position}
Juan~M Escano, Manuel~G Ortega, and Francisco~R Rubio.
\newblock Position control of a pneumatic levitation system.
\newblock In \emph{2005 IEEE Conference on Emerging Technologies and Factory
  Automation}, volume~1, pages 6--pp. IEEE, 2005.
\newblock URL
  \url{https://ieeexplore.ieee.org/stamp/stamp.jsp?tp=&arnumber=1612568}.

\bibitem[Ganapathi et~al.(2021)Ganapathi, Sundaresan, Thananjeyan, Balakrishna,
  Seita, Grannen, Hwang, Hoque, Gonzalez, Jamali,
  et~al.]{ganapathi2021learning}
Aditya Ganapathi, Priya Sundaresan, Brijen Thananjeyan, Ashwin Balakrishna,
  Daniel Seita, Jennifer Grannen, Minho Hwang, Ryan Hoque, Joseph~E Gonzalez,
  Nawid Jamali, et~al.
\newblock Learning dense visual correspondences in simulation to smooth and
  fold real fabrics.
\newblock In \emph{2021 IEEE International Conference on Robotics and
  Automation (ICRA)}, pages 11515--11522. IEEE, 2021.

\bibitem[Ha and Song(2021)]{ha2021flingbot}
Huy Ha and Shuran Song.
\newblock Flingbot: The unreasonable effectiveness of dynamic manipulation for
  cloth unfolding.
\newblock In \emph{Conference on Robotic Learning (CoRL)}, 2021.
\newblock URL \url{https://arxiv.org/pdf/2105.03655.pdf}.

\bibitem[Ha et~al.(2020)Ha, Xu, and Song]{ha2020learning}
Huy Ha, Jingxi Xu, and Shuran Song.
\newblock Learning a decentralized multi-arm motion planner.
\newblock \emph{CoRL}, 2020.

\bibitem[Howard and Bekey(1996)]{howard1996prototype}
Ayanna~M Howard and George~A Bekey.
\newblock Prototype system for automated sorting and removal of bags of
  hazardous waste.
\newblock In \emph{Intelligent Robots and Computer Vision XV: Algorithms,
  Techniques, Active Vision, and Materials Handling}, volume 2904, pages
  271--277. International Society for Optics and Photonics, 1996.

\bibitem[Howard and Bekey(2000)]{howard2000intelligent}
Ayanna~M Howard and George~A Bekey.
\newblock Intelligent learning for deformable object manipulation.
\newblock \emph{Autonomous Robots}, 9\penalty0 (1):\penalty0 51--58, 2000.

\bibitem[Jangir et~al.(2019)Jangir, Alenya, and Torras]{jangir2019dynamic}
Rishabh Jangir, Guillem Alenya, and Carme Torras.
\newblock Dynamic cloth manipulation with deep reinforcement learning.
\newblock \emph{arXiv preprint arXiv:1910.14475}, 2019.

\bibitem[Konishi and Fujita(1994)]{konishi1994conveyance}
Satoshi Konishi and Hiroyuki Fujita.
\newblock A conveyance system using air flow based on the concept of
  distributed micro motion systems.
\newblock \emph{Journal of microelectromechanical systems}, 3\penalty0
  (2):\penalty0 54--58, 1994.

\bibitem[Konishi and Fujita(1996)]{konishi1996two}
Satoshi Konishi and Hiroyuki Fujita.
\newblock Two-dimensional conveyance system using cooperative motions of many
  microactuators.
\newblock In \emph{Proceedings of IEEE/RSJ International Conference on
  Intelligent Robots and Systems. IROS'96}, volume~2, pages 988--992. IEEE,
  1996.

\bibitem[Konishi et~al.(1999)Konishi, Mizuguchi, and
  Ohno]{konishi1999development}
Satoshi Konishi, Yasushi Mizuguchi, and Kazuyuki Ohno.
\newblock Development of a non-contact conveyance system composed of
  distributed nozzle units.
\newblock In \emph{1999 7th IEEE International Conference on Emerging
  Technologies and Factory Automation. Proceedings ETFA'99 (Cat. No.
  99TH8467)}, volume~1, pages 593--598. IEEE, 1999.
\newblock URL
  \url{https://ieeexplore.ieee.org/stamp/stamp.jsp?tp=&arnumber=815409}.

\bibitem[Lee et~al.(2020)Lee, Ward, Cosgun, Dasagi, Corke, and
  Leitner]{lee2020learning}
Robert Lee, Daniel Ward, Akansel Cosgun, Vibhavari Dasagi, Peter Corke, and
  Jurgen Leitner.
\newblock Learning arbitrary-goal fabric folding with one hour of real robot
  experience.
\newblock In \emph{Conference on Robotic Learning (CoRL)}, 2020.
\newblock URL \url{https://arxiv.org/pdf/2010.03209.pdf}.

\bibitem[Li et~al.(2019)Li, Wu, Tedrake, Tenenbaum, and
  Torralba]{li2018learning}
Yunzhu Li, Jiajun Wu, Russ Tedrake, Joshua~B Tenenbaum, and Antonio Torralba.
\newblock Learning particle dynamics for manipulating rigid bodies, deformable
  objects, and fluids.
\newblock In \emph{ICLR}, 2019.
\newblock URL \url{http://dpi.csail.mit.edu/dpi-paper.pdf}.

\bibitem[Lim et~al.(2021)Lim, Huang, Chen, Wang, Ichnowski, Seita, Laskey, and
  Goldberg]{casting}
Vincent Lim, Huang Huang, Lawrence~Yunliang Chen, Jonathan Wang, Jeffrey
  Ichnowski, Daniel Seita, Michael Laskey, and Ken Goldberg.
\newblock Planar robot casting with real2sim2real self-supervised learning.
\newblock \emph{arXiv preprint arXiv:2111.04814}, 2021.

\bibitem[Lin et~al.(2020)Lin, Wang, Olkin, and Held]{lin2020softgym}
Xingyu Lin, Yufei Wang, Jake Olkin, and David Held.
\newblock Softgym: Benchmarking deep reinforcement learning for deformable
  object manipulation.
\newblock \emph{CoRL}, 2020.

\bibitem[Lin et~al.(2021)Lin, Wang, Huang, and Held]{lin2021VCD}
Xingyu Lin, Yufei Wang, Zixuan Huang, and David Held.
\newblock Learning visible connectivity dynamics for cloth smoothing.
\newblock In \emph{Conference on Robot Learning}, 2021.

\bibitem[Maitin-Shepard et~al.(2010)Maitin-Shepard, Cusumano-Towner, Lei, and
  Abbeel]{maitin2010cloth}
Jeremy Maitin-Shepard, Marco Cusumano-Towner, Jinna Lei, and Pieter Abbeel.
\newblock Cloth grasp point detection based on multiple-view geometric cues
  with application to robotic towel folding.
\newblock In \emph{2010 IEEE International Conference on Robotics and
  Automation}, pages 2308--2315. IEEE, 2010.

\bibitem[Mason and Lynch(1993)]{mason1993dynamic}
Matthew~T Mason and Kevin~M Lynch.
\newblock Dynamic manipulation.
\newblock In \emph{Proceedings of 1993 IEEE/RSJ International Conference on
  Intelligent Robots and Systems (IROS'93)}, volume~1, pages 152--159. IEEE,
  1993.
\newblock URL
  \url{http://citeseerx.ist.psu.edu/viewdoc/download?doi=10.1.1.15.1943&rep=rep1&type=pdf}.

\bibitem[Nordine and Atkins(1982)]{nordine1982aerodynamic}
Paul~C Nordine and Robert~M Atkins.
\newblock Aerodynamic levitation of laser-heated solids in gas jets.
\newblock \emph{Review of Scientific Instruments}, 53\penalty0 (9):\penalty0
  1456--1464, 1982.
\newblock URL \url{https://aip.scitation.org/doi/pdf/10.1063/1.1137196}.

\bibitem[Ozcelik and Erzincanli(2002)]{ozcelik2002non}
Babur Ozcelik and Fehmi Erzincanli.
\newblock A non-contact end-effector for the handling of garments.
\newblock \emph{Robotica}, 20\penalty0 (4):\penalty0 447--450, 2002.
\newblock URL
  \url{https://www.cambridge.org/core/journals/robotica/article/noncontact-endeffector-for-the-handling-of-garments/20711AA699FAA3665F34030BEDC95E49}.

\bibitem[Ozcelik and Erzincanli(2005)]{ozcelik2005examination}
Babur Ozcelik and Fehmi Erzincanli.
\newblock Examination of the movement of a woven fabric in the horizontal
  direction using a non-contact end-effector.
\newblock \emph{The International Journal of Advanced Manufacturing
  Technology}, 25\penalty0 (5):\penalty0 527--532, 2005.
\newblock URL
  \url{https://link.springer.com/article/10.1007/s00170-004-2075-x}.

\bibitem[Paszke et~al.(2019)Paszke, Gross, Massa, Lerer, Bradbury, Chanan,
  Killeen, Lin, Gimelshein, Antiga, et~al.]{paszke2019pytorch}
Adam Paszke, Sam Gross, Francisco Massa, Adam Lerer, James Bradbury, Gregory
  Chanan, Trevor Killeen, Zeming Lin, Natalia Gimelshein, Luca Antiga, et~al.
\newblock Pytorch: An imperative style, high-performance deep learning library.
\newblock \emph{Advances in neural information processing systems},
  32:\penalty0 8026--8037, 2019.
\newblock URL \url{https://arxiv.org/pdf/1912.01703.pdf}.

\bibitem[Pister et~al.(1990)Pister, Fearing, and Howe]{pister1990planar}
Kristopher~SJ Pister, Ronald~S Fearing, and Roger~T Howe.
\newblock A planar air levitated electrostatic actuator system.
\newblock In \emph{IEEE Proceedings on Micro Electro Mechanical Systems, An
  Investigation of Micro Structures, Sensors, Actuators, Machines and Robots.},
  pages 67--71. IEEE, 1990.

\bibitem[Rus and Tolley(2015)]{rus2015design}
Daniela Rus and Michael~T Tolley.
\newblock Design, fabrication and control of soft robots.
\newblock \emph{Nature}, 521\penalty0 (7553):\penalty0 467--475, 2015.

\bibitem[Sanan et~al.(2009)Sanan, Moidel, and Atkeson]{sanan2009robots}
Siddharth Sanan, Justin~B Moidel, and Christopher~G Atkeson.
\newblock Robots with inflatable links.
\newblock In \emph{2009 IEEE/RSJ International Conference on Intelligent Robots
  and Systems}, pages 4331--4336. IEEE, 2009.
\newblock URL
  \url{https://ieeexplore.ieee.org/stamp/stamp.jsp?tp=&arnumber=5354151}.

\bibitem[Seita et~al.(2018)Seita, Jamali, Laskey, Berenstein, Tanwani,
  Baskaran, Iba, Canny, and Goldberg]{seita2018bedmaking}
Daniel Seita, Nawid Jamali, Michael Laskey, Ron Berenstein, Ajay~Kumar Tanwani,
  Prakash Baskaran, Soshi Iba, John~F. Canny, and Ken Goldberg.
\newblock Robot bed-making: Deep transfer learning using depth sensing of
  deformable fabric.
\newblock \emph{CoRR}, abs/1809.09810, 2018.
\newblock URL \url{http://arxiv.org/abs/1809.09810}.

\bibitem[Seita et~al.(2020)Seita, Ganapathi, Hoque, Hwang, Cen, Tanwani,
  Balakrishna, Thananjeyan, Ichnowski, Jamali, et~al.]{seita2020deep}
Daniel Seita, Aditya Ganapathi, Ryan Hoque, Minho Hwang, Edward Cen, Ajay~Kumar
  Tanwani, Ashwin Balakrishna, Brijen Thananjeyan, Jeffrey Ichnowski, Nawid
  Jamali, et~al.
\newblock Deep imitation learning of sequential fabric smoothing from an
  algorithmic supervisor.
\newblock In \emph{2020 IEEE/RSJ International Conference on Intelligent Robots
  and Systems (IROS)}, pages 9651--9658. IEEE, 2020.

\bibitem[Seita et~al.(2021{\natexlab{a}})Seita, Florence, Tompson, Coumans,
  Sindhwani, Goldberg, and Zeng]{seita2021learning}
Daniel Seita, Pete Florence, Jonathan Tompson, Erwin Coumans, Vikas Sindhwani,
  Ken Goldberg, and Andy Zeng.
\newblock Learning to rearrange deformable cables, fabrics, and bags with
  goal-conditioned transporter networks.
\newblock In \emph{2021 IEEE International Conference on Robotics and
  Automation (ICRA)}, pages 4568--4575. IEEE, 2021{\natexlab{a}}.

\bibitem[Seita et~al.(2021{\natexlab{b}})Seita, Kerr, Canny, and
  Goldberg]{seita_bags_iros_2021}
Daniel Seita, Justin Kerr, John Canny, and Ken Goldberg.
\newblock {Initial Results on Grasping and Lifting Physical Deformable Bags
  with a Bimanual Robot}.
\newblock In \emph{IROS Workshop on Robotic Manipulation of Deformable Objects
  in Real-world Applications}, 2021{\natexlab{b}}.

\bibitem[Shibata et~al.(2010)Shibata, Ohta, and Hirai]{shibata2010robotic}
Mizuho Shibata, Tsuyoshi Ohta, and Shinichi Hirai.
\newblock Robotic unfolding of hemmed fabric using pinching slip motion.
\newblock In \emph{The Abstracts of the international conference on advanced
  mechatronics: toward evolutionary fusion of IT and mechatronics: ICAM
  2010.5}, pages 392--397. The Japan Society of Mechanical Engineers, 2010.

\bibitem[Sundaresan et~al.(2020)Sundaresan, Grannen, Thananjeyan, Balakrishna,
  Laskey, Stone, Gonzalez, and Goldberg]{sundaresan2020learning}
Priya Sundaresan, Jennifer Grannen, Brijen Thananjeyan, Ashwin Balakrishna,
  Michael Laskey, Kevin Stone, Joseph~E Gonzalez, and Ken Goldberg.
\newblock Learning rope manipulation policies using dense object descriptors
  trained on synthetic depth data.
\newblock In \emph{2020 IEEE International Conference on Robotics and
  Automation (ICRA)}, pages 9411--9418. IEEE, 2020.

\bibitem[Tootchi et~al.(2019)Tootchi, Amirkhani, and
  Chaibakhsh]{tootchi2019modeling}
Amirreza Tootchi, Saeed Amirkhani, and Ali Chaibakhsh.
\newblock Modeling and control of an air levitation ball and pipe laboratory
  setup.
\newblock In \emph{2019 7th International Conference on Robotics and
  Mechatronics (ICRoM)}, pages 29--34. IEEE, 2019.
\newblock URL
  \url{https://ieeexplore.ieee.org/stamp/stamp.jsp?tp=&arnumber=9071827}.

\bibitem[Wang et~al.(2020)Wang, Wang, Romero, Veiga, and
  Adelson]{wang2020swingbot}
Chen Wang, Shaoxiong Wang, Branden Romero, Filipe Veiga, and Edward Adelson.
\newblock Swingbot: Learning physical features from in-hand tactile exploration
  for dynamic swing-up manipulation.
\newblock In \emph{2020 IEEE/RSJ International Conference on Intelligent Robots
  and Systems (IROS)}, pages 5633--5640. IEEE, 2020.
\newblock URL \url{https://arxiv.org/pdf/2101.11812.pdf}.

\bibitem[Wasbari et~al.(2017)Wasbari, Bakar, Gan, Tahir, and
  Yusof]{wasbari2017review}
Faizil Wasbari, RA~Bakar, Leong~Ming Gan, MM~Tahir, and Ahmad~Anas Yusof.
\newblock A review of compressed-air hybrid technology in vehicle system.
\newblock \emph{Renewable and Sustainable Energy Reviews}, 67:\penalty0
  935--953, 2017.

\bibitem[Weng et~al.(2021{\natexlab{a}})Weng, Bajracharya, Wang, Agrawal, and
  Held]{weng2021fabricflownet}
Thomas Weng, Sujay Bajracharya, Yufei Wang, Khush Agrawal, and David Held.
\newblock Fabricflownet: Bimanual cloth manipulation with a flow-based policy.
\newblock In \emph{Conference on Robot Learning}, 2021{\natexlab{a}}.

\bibitem[Weng et~al.(2021{\natexlab{b}})Weng, Paus, Varava, Yin, Asfour, and
  Kragic]{weng2021graph}
Zehang Weng, Fabian Paus, Anastasiia Varava, Hang Yin, Tamim Asfour, and Danica
  Kragic.
\newblock Graph-based task-specific prediction models for interactions between
  deformable and rigid objects.
\newblock \emph{arXiv preprint arXiv:2103.02932}, 2021{\natexlab{b}}.

\bibitem[Wu et~al.(2020)Wu, Sun, Zeng, Song, Lee, Rusinkiewicz, and
  Funkhouser]{wu2020spatial}
Jimmy Wu, Xingyuan Sun, Andy Zeng, Shuran Song, Johnny Lee, Szymon
  Rusinkiewicz, and Thomas Funkhouser.
\newblock Spatial action maps for mobile manipulation.
\newblock In \emph{Proceedings of Robotics: Science and Systems (RSS)}, 2020.
\newblock URL \url{https://arxiv.org/abs/2004.09141}.

\bibitem[Wu et~al.(2022)Wu, Sun, Zeng, Song, Rusinkiewicz, and
  Funkhouser]{wu2022learning}
Jimmy Wu, Xingyuan Sun, Andy Zeng, Shuran Song, Szymon Rusinkiewicz, and Thomas
  Funkhouser.
\newblock Learning pneumatic non-prehensile manipulation with a mobile blower.
\newblock \emph{arXiv preprint arXiv:2204.02390}, 2022.

\bibitem[Wu et~al.(2019)Wu, Yan, Kurutach, Pinto, and Abbeel]{wu2019learning}
Yilin Wu, Wilson Yan, Thanard Kurutach, Lerrel Pinto, and Pieter Abbeel.
\newblock Learning to manipulate deformable objects without demonstrations.
\newblock \emph{arXiv preprint arXiv:1910.13439}, 2019.

\bibitem[Xu et~al.(2019)Xu, Wu, Zeng, Tenenbaum, and Song]{DensePhysNet}
Zhenjia Xu, Jiajun Wu, Andy Zeng, Joshua~B Tenenbaum, and Shuran Song.
\newblock Densephysnet: Learning dense physical object representations via
  multi-step dynamic interactions.
\newblock In \emph{Robotics: Science and Systems (RSS)}, 2019.

\bibitem[Xu et~al.(2022)Xu, Zhanpeng, and Song]{xu2022umpnet}
Zhenjia Xu, He~Zhanpeng, and Shuran Song.
\newblock Umpnet: Universal manipulation policy network for articulated
  objects.
\newblock \emph{IEEE Robotics and Automation Letters}, 2022.

\bibitem[Yamaguchi et~al.(2013)Yamaguchi, Hirata, and
  Kosuge]{yamaguchi2013development}
Kengo Yamaguchi, Yasuhisa Hirata, and Kazuhiro Kosuge.
\newblock Development of robot hand with suction mechanism for robust and
  dexterous grasping.
\newblock In \emph{2013 IEEE/RSJ International Conference on Intelligent Robots
  and Systems}, pages 5500--5505. IEEE, 2013.

\bibitem[Yamakawa et~al.(2011)Yamakawa, Namiki, and
  Ishikawa]{yamakawa2011dynamic}
Yuji Yamakawa, Akio Namiki, and Masatoshi Ishikawa.
\newblock Dynamic manipulation of a cloth by high-speed robot system using
  high-speed visual feedback.
\newblock \emph{IFAC Proceedings Volumes}, 44\penalty0 (1):\penalty0
  8076--8081, 2011.

\bibitem[Zeng et~al.(2018)Zeng, Song, Welker, Lee, Rodriguez, and
  Funkhouser]{zeng2018learning}
Andy Zeng, Shuran Song, Stefan Welker, Johnny Lee, Alberto Rodriguez, and
  Thomas Funkhouser.
\newblock Learning synergies between pushing and grasping with self-supervised
  deep reinforcement learning.
\newblock In \emph{Proceedings of the IEEE International Conference on
  Intelligent Robots and Systems (IROS)}, 2018.
\newblock URL \url{https://arxiv.org/abs/1803.09956}.

\bibitem[Zeng et~al.(2020)Zeng, Song, Lee, Rodriguez, and
  Funkhouser]{zeng2020tossingbot}
Andy Zeng, Shuran Song, Johnny Lee, Alberto Rodriguez, and Thomas Funkhouser.
\newblock Tossingbot: Learning to throw arbitrary objects with residual
  physics.
\newblock \emph{IEEE Transactions on Robotics}, 36\penalty0 (4):\penalty0
  1307--1319, 2020.
\newblock URL
  \url{https://ieeexplore.ieee.org/stamp/stamp.jsp?arnumber=9104757}.

\bibitem[Zhakypov et~al.(2018)Zhakypov, Heremans, Billard, and
  Paik]{zhakypov2018origami}
Zhenishbek Zhakypov, Florian Heremans, Aude Billard, and Jamie Paik.
\newblock An origami-inspired reconfigurable suction gripper for picking
  objects with variable shape and size.
\newblock \emph{IEEE Robotics and Automation Letters}, 3\penalty0 (4):\penalty0
  2894--2901, 2018.
\newblock URL
  \url{https://ieeexplore.ieee.org/stamp/stamp.jsp?tp=&arnumber=8385192}.

\bibitem[Zhang et~al.(2021)Zhang, Ichnowski, Seita, Wang, Huang, and
  Goldberg]{zhang2021robots}
Harry Zhang, Jeffrey Ichnowski, Daniel Seita, Jonathan Wang, Huang Huang, and
  Ken Goldberg.
\newblock Robots of the lost arc: Self-supervised learning to dynamically
  manipulate fixed-endpoint cables.
\newblock In \emph{2021 IEEE International Conference on Robotics and
  Automation (ICRA)}, pages 4560--4567. IEEE, 2021.

\end{thebibliography}

\begin{figure*}[!t]
    \centering
    \includegraphics[width=\linewidth]{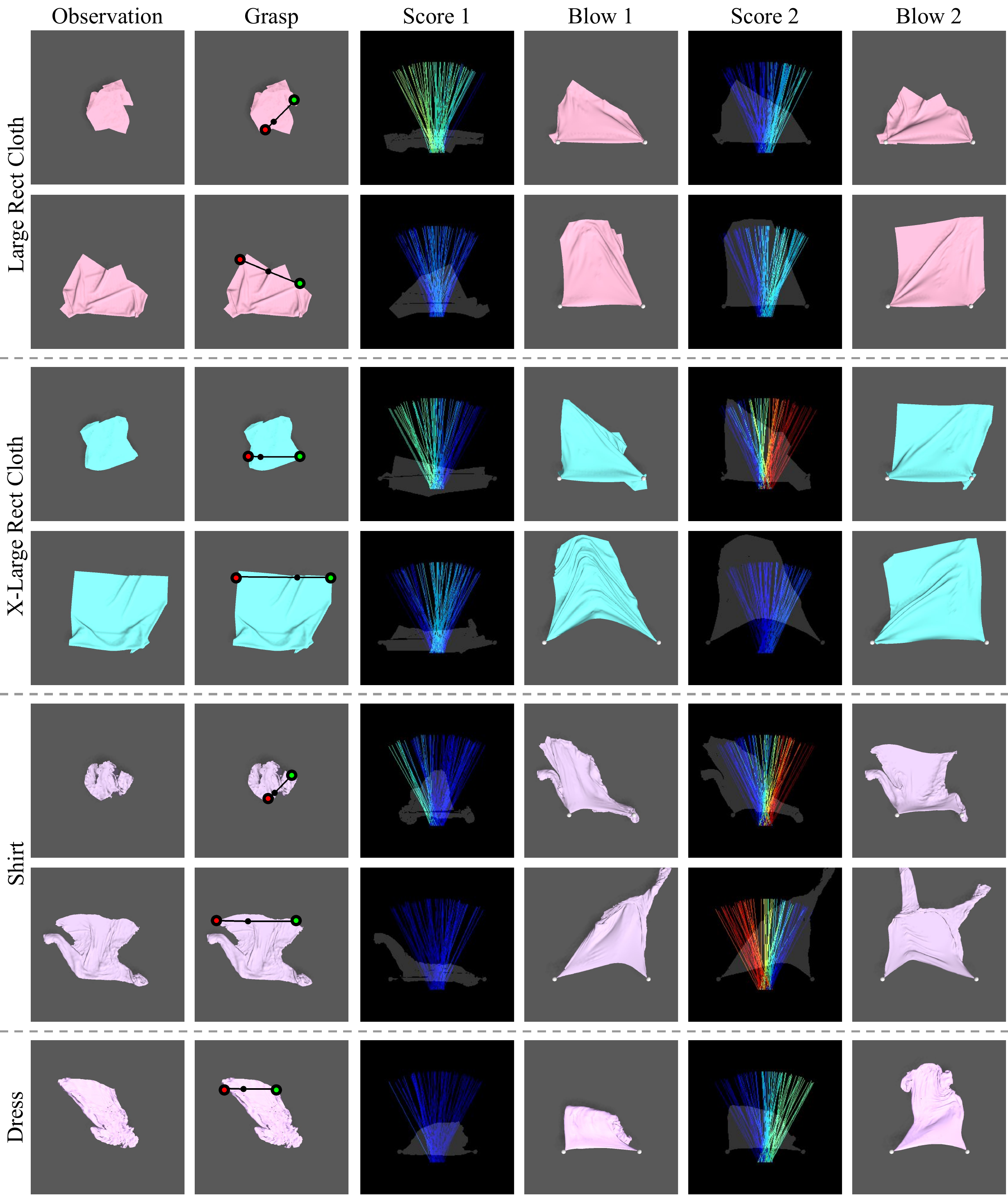}
    \caption{\textbf{Additional cloth unfolding results.}}
    \label{fig:supp-cloth}
\end{figure*}
\begin{figure*}[!t]
    \centering
    \includegraphics[width=\linewidth]{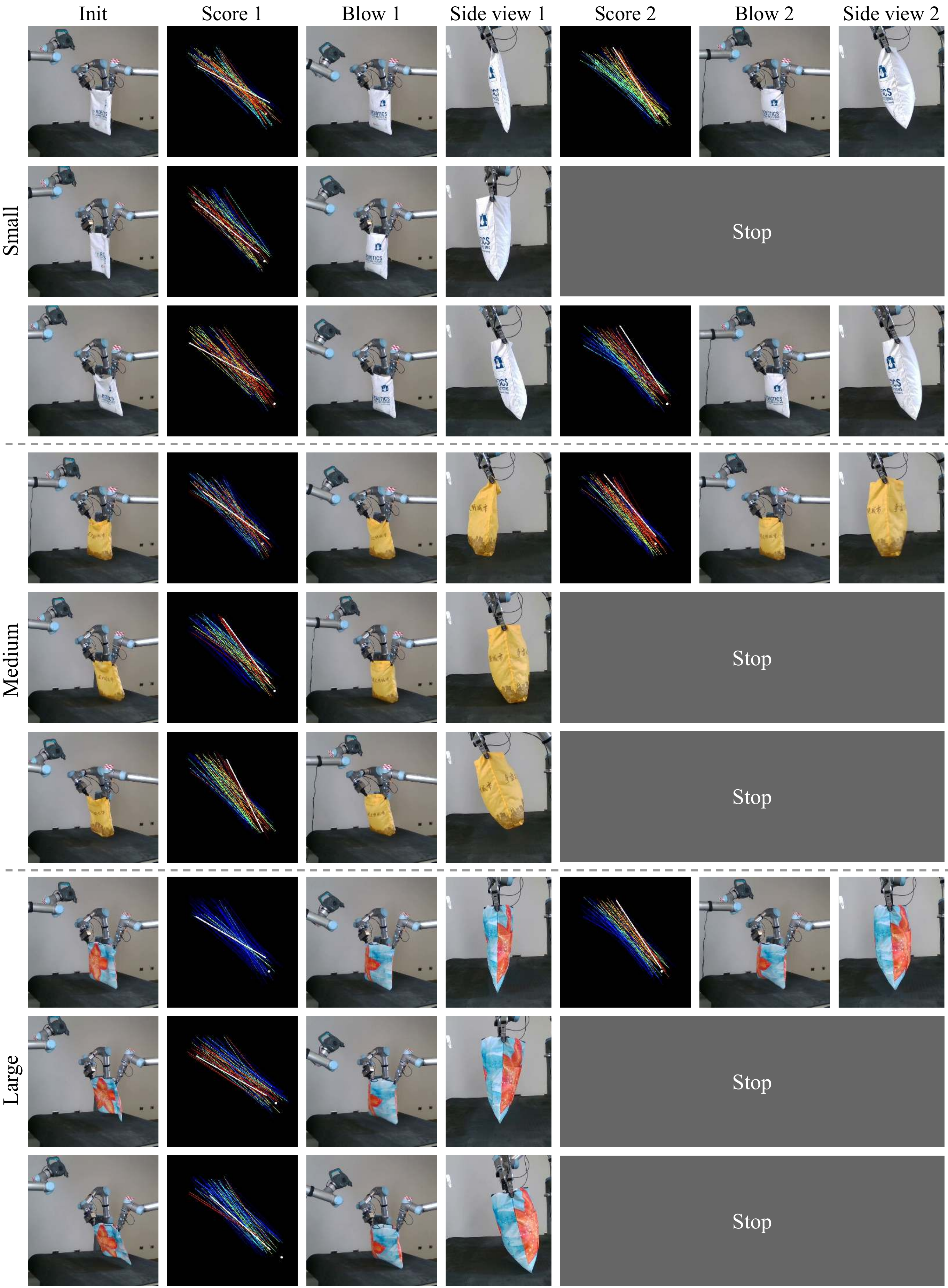}
    \caption{\textbf{Additional Bag opening results.}}
    \label{fig:supp-bag}
\end{figure*}

\end{document}